\documentclass[11pt]{article}

\usepackage[preprint]{acl}

\usepackage{times}
\usepackage{latexsym}
\usepackage{amsmath,amsfonts}

\usepackage[T1]{fontenc}
\usepackage{booktabs}
\usepackage{multirow}
\usepackage{graphicx}

\usepackage[utf8]{inputenc}

\usepackage{microtype}

\usepackage{inconsolata}

\usepackage{graphicx}
\usepackage{subcaption}
\usepackage{amsmath}
\usepackage{todonotes}
\usepackage{multirow}

\title{Language Models are not Equally Robust to\\ Non-Canonical Tokenization across Languages}

\author{Poulami Ghosh$^{\ast}$, Preethi Jyothi$^{\ast}$\\ { }
$^{\ast}$IIT Bombay, India\\
\texttt{\{poulami, pjyothi}\}@cse.iitb.ac.in}

\begin{document}
\maketitle
\begin{abstract}
Despite the existence of exponentially many valid tokenizations for a given string, language models operate on a single canonical sequence deterministically produced by the tokenizer, leaving the broader tokenization space largely uncharacterized. In this paper, we investigate this overlooked space by studying the behavior of language models under non-canonical tokenizations across diverse languages. For English, prior work shows that models are largely invariant to alternative tokenizations that represent the same underlying string. We ask whether this invariance generalizes to other languages beyond English. We conduct a multilingual study across 27 languages spanning diverse scripts and evaluate LLM behavior under alternative tokenizations across six downstream tasks. We find that tokenization invariance does not generalize: model behavior varies substantially across languages with instruction-tuned models exhibiting an average relative performance drop of 23.7\% for Llama-3.1-8B, 11.4\% for Qwen3-8B, and 9.9\% for Gemma-3-12B.
The variation of tokenization invariance is systematic across languages. Languages that exhibit higher token fragmentation show significantly greater sensitivity to non-canonical tokenizations.
Our study of tokenization robustness serves as a diagnostic of how tightly a model is coupled to its tokenizer. These results demonstrate that tokenization robustness is not a universal property of language models, but depends strongly on the language and its interaction with the tokenizer. We show that LoRA fine-tuning with multi-tokenization training data provides an effective mitigation for tokenization sensitivity. Fine-tuning on English alone improves tokenization robustness across languages, while systematically sampling diverse non-canonical tokenizations achieves the strongest overall performance.
\end{abstract}

\section{Introduction}
Tokenization plays a central role in modern language models, determining how raw text is segmented into the discrete units processed by the model. Most contemporary systems rely on subword tokenization algorithms such as \citet{sennrich-etal-2016-neural} (Byte Pair Encoding), \citet{kudo-richardson-2018-sentencepiece} (SentencePiece), \citet{wu2016google} (WordPiece), \citet{kudo-2018-subword} (UnigramLM). Although tokenization is often treated as a preprocessing step, it directly shapes the model’s input distribution and therefore the representations learned during training. The tokenizer typically produces a canonical tokenization, a deterministic segmentation of the input text according to its tokenization algorithm, which defines the input distribution seen during pretraining. However, the same text can often be represented using many alternative segmentations composed entirely of valid vocabulary tokens. These non-canonical tokenizations preserve the surface text while altering only the sequence of input tokens. Since contemporary LLMs are trained almost exclusively on canonical segmentations, such inputs lie outside the training distribution. Consequently, one would generally expect degradation in model performance under such perturbation.
Surprisingly, however, \cite{zheng2025broken} found that modern instruction-tuned LLMs remain remarkably robust to non-canonical tokenizations in English across a range of downstream tasks. They hypothesize that this robustness emerges during post-training, where models are encouraged to respond fluently regardless of variations in the tokenization of user instructions.


\begin{figure*}[t!]
	\centering
	\includegraphics[width=0.99\textwidth]{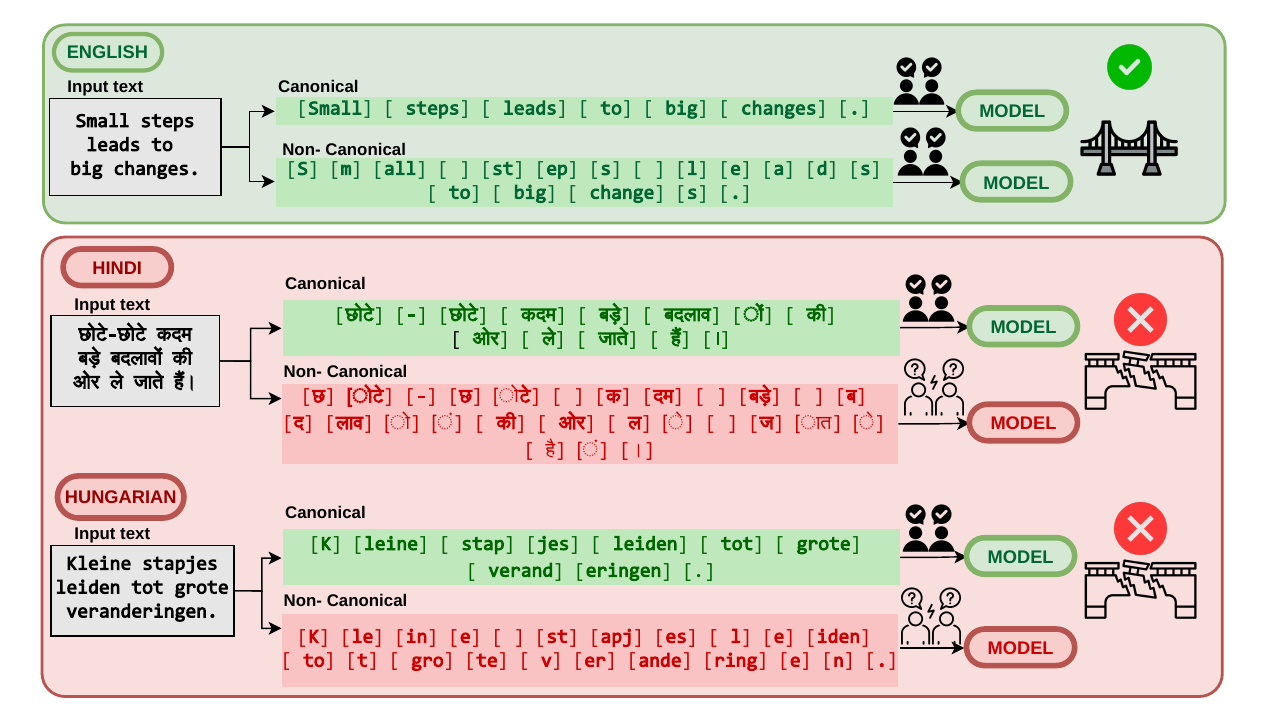}
	\caption{LLM robustness to non-canonical tokenization is \textbf{NOT} universal and varies substantially across languages, models, and tasks. Models are "broken" in multilingual settings.} 
    \label{fig:overview}
\end{figure*}

Whether this surprising robustness extends beyond English remains unclear. English occupies a privileged position in multilingual tokenizers: it is heavily represented during tokenizer construction, exhibits relatively simple morphology, and benefits from whitespace-delimited segmentation. In contrast, multilingual tokenizers must allocate a fixed vocabulary across many languages with diverse scripts and morphological structures. 
Prior work has shown that vocabulary allocation is highly uneven across languages \cite{limisiewicz-etal-2023-tokenization,petrov2023language,ahia2023all}, particularly disadvantaging lower-resource languages. Taken together, these factors suggest that robustness findings derived from English cannot be assumed to generalize to other languages.

Although non-canonical tokenizations lie outside the training distribution, they are not merely synthetic perturbations. They naturally arise in several practical settings where the tokenizer is modified after pretraining. Recent work on vocabulary pruning and restructuring \cite{reif-etal-2026-vocab,purason-etal-2026-teaching,kim2026optimizing} aims to improve efficiency and language coverage by removing redundant tokens or reallocating vocabulary capacity, causing the same text to be segmented differently after tokenizer adaptation. Similarly, cross-lingual tokenizer transfer methods replace or adapt the original tokenizer to better support new languages or domains through through embedding transfer or token alignment \cite{remy2024trans,li2025tokalign,li2026tokalign++}. Such tokenizer replacement can also change how the same input would originally get segmented. 
Beyond tokenizer adaptation, \citet{geh2025adversarial} show that adversarially selected non-canonical tokenizations of malicious strings can bypass LLM safety and alignment mechanisms while preserving the underlying text, although this has only been demonstrated for English. These demonstrate practical scenarios where a model may encounter non-canonical tokenization during inference and motivate understanding how robust multilingual language models are to alternative valid tokenizations.


In this work, we systematically study how multilingual language models respond to non-canonical tokenization across languages. Rather than degrading uniformly, multilingual LLMs exhibit strikingly different levels of robustness to non-canonical tokenization across models, languages, and tasks. Our findings reveal that tokenization robustness is not simply a by-product of stronger language models, but a distinct property of multilingual language models shaped by language-specific tokenization characteristics such as token fragmentation. Moreover, the effects extend beyond text generation. Even purely semantic tasks like cross-lingual sentence retrieval that rely on sentence-level embeddings are significantly impacted by non-canonical tokenizations. 
Our contributions are:\\
\noindent\textbf{1.} We present the first systematic study of non-canonical tokenization in multilingual LLMs, spanning 27 languages, six downstream tasks, and three recent model families.\\
\noindent\textbf{2.} We analyze how non-canonical tokenization affects multilingual language models differently across language families and scripts.\\
\noindent\textbf{3.} We show that non-canonical tokenization serves as a natural source of data augmentation for improving multilingual generalization.\\

\section{Related Work} 
\paragraph{Robustness to Non-Canonical Tokenization.}
Recent work challenges the assumption that language models critically depend on a fixed canonical tokenization. \citet{zheng2025broken} show that modern LLMs retain strong performance even under heavily perturbed 
non-canonical tokenizations, suggesting substantial robustness at inference time. Similarly, \citet{geh2025adversarial} demonstrate that adversarial tokenizations
can significantly alter model behavior, including bypassing alignment safeguards, without changing the underlying text. Together, these works establish that tokenization is not merely a preprocessing step but directly influences inference. However, both studies are primarily conducted in English and do not examine how such robustness or sensitivity varies across languages with different scripts. 

\paragraph{Tokenization as a Latent Probabilistic Space.}
A complementary line of work treats tokenization as a source of latent variability in language modeling. \citet{cao2021you} and \citet{chirkova2023should} argue that evaluation should account for multiple tokenizations to better approximate the true likelihood of a sequence. 
\citet{geh2024signal} introduces the idea that tokenization space contains latent signal beyond the canonical sequence. They show that aggregating probabilities across tokenizations can uncover useful signal and improve evaluation performance. While these works highlight the importance of tokenization beyond a single canonical segmentation, they focus primarily on likelihood estimation rather than its effect on generative downstream performance, particularly in multilingual settings.


\paragraph{Learning Under Tokenization Variability.}
Prior work has also explored exposing models to multiple tokenizations during training. \citet{cognetta2024distributional} analyze subword regularization, showing that sampling alternative segmentations improves robustness and generalization. \citet{steger2026stochasticity} demonstrated that training with stochastically sampled non-canonical tokenizations improves robustness to both random and adversarial tokenization perturbations, with uniform sampling strategies providing stronger robustness while preserving canonical performance. At the extreme, \citet{vieira2024language} eliminate tokenization entirely by modeling text at the character level. These approaches suggest that models can benefit from tokenization variability, but they focus on training-time interventions or alternative architectures, rather than inference-time manipulation of tokenization in pretrained LLMs.

However, all these directions remain largely disconnected and predominantly English-centric. In particular, there is limited understanding of how non-canonical tokenization affects inference across languages from different language families with diverse scripts. 
Our work addresses this gap by systematically studying the impact of non-canonical tokenization on LLM inference in a multilingual setting.

\section{Methodology}
Let $\Sigma$ be a finite alphabet and $\Sigma^*$ the set of all finite strings over $\Sigma$. Let $V$ denote a tokenizer vocabulary, and $V^*$ the set of finite token sequences.

\paragraph{Tokenization and decoding.}
A tokenizer induces a \emph{deterministic canonical tokenization}
\[
T_{\mathrm{canon}} : \Sigma^* \rightarrow V^*, 
\quad
t_{\mathrm{canon}} = T_{\mathrm{canon}}(x),
\]
which maps any string $x \in \Sigma^*$ to a unique token sequence.

A \emph{decoding function}
$
\mathrm{decode} : V^* \rightarrow \Sigma^*
$
maps token sequences back to strings, satisfying
\[
\mathrm{decode}(T_{\mathrm{canon}}(x)) = x.
\]

We define the set of all valid tokenizations of a string $x$ as:
\[
\mathcal{T}(x) = \{\, t \in V^* \;:\; \mathrm{decode}(t) = x \,\}.
\]
By construction, $
T_{\mathrm{canon}}(x) \in \mathcal{T}(x)
$.

\paragraph{Tokenization invariance.}
Let a language model be a function
\[
f : V^* \rightarrow \mathbb{R}^d,
\]
mapping token sequences to outputs (e.g., logits, representations, or task predictions). We say the model satisfies \emph{tokenization invariance} if for tokenizations $
t_1, t_2 \in \mathcal{T}(x)
$, 
\[
f(t_1) \approx f(t_2).
\]
This captures the ideal requirement that model behavior should depend on the underlying string
x, not on how it is segmented into tokens.
Given two tokenizations $t_1, t_2 \in \mathcal{T}(x)$, we quantify the deviation from tokenization invariance as:
\[
\mathrm{Err}(t_1, t_2) = d\big(f(t_1), f(t_2)\big),
\]
where $d$ 
measures tokenization-induced performance gap. In our evaluation setup, we compare a reference canonical tokenization with different non-canonical tokenizations.

\begin{figure*}[t!]
\centering
\begin{subfigure}{0.23\textwidth}
    \centering
    \includegraphics[width=\linewidth]{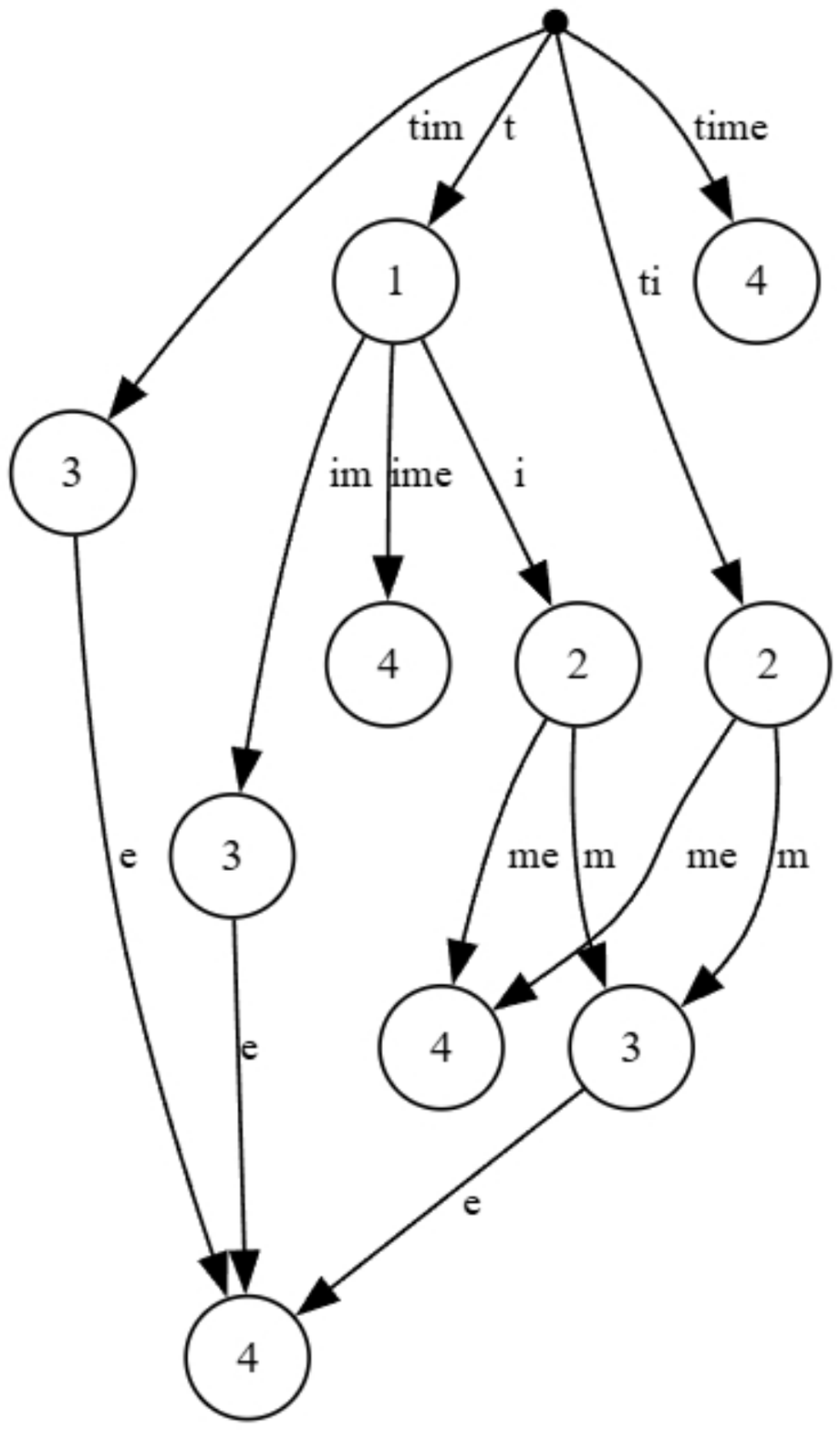}
    \caption{English}
\end{subfigure}
\hfill
\begin{subfigure}{0.23\textwidth}
    \centering
    \includegraphics[width=\linewidth]{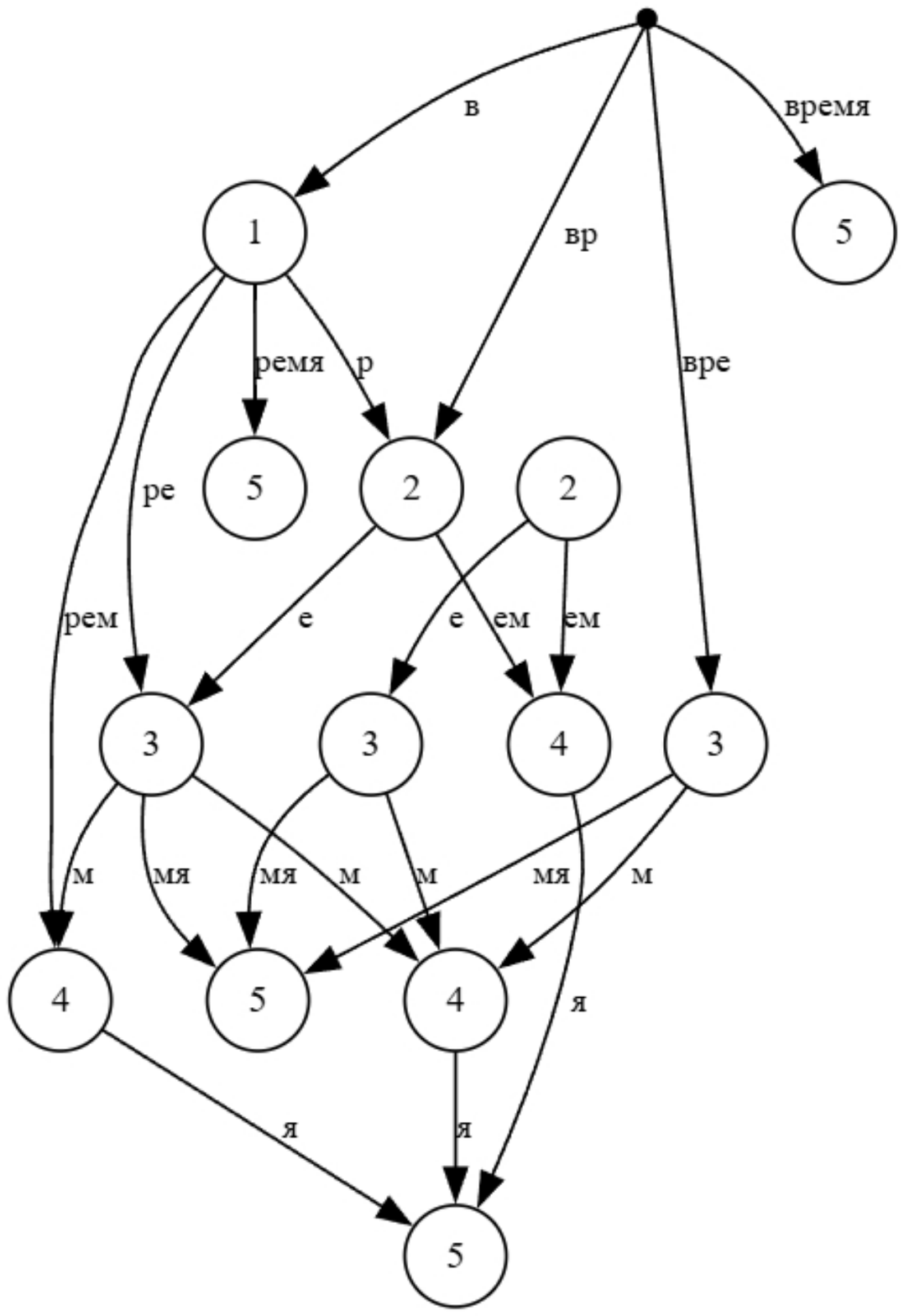}
    \caption{Russian}
\end{subfigure}
\hfill
\begin{subfigure}{0.23\textwidth}
    \centering
    \includegraphics[width=\linewidth]{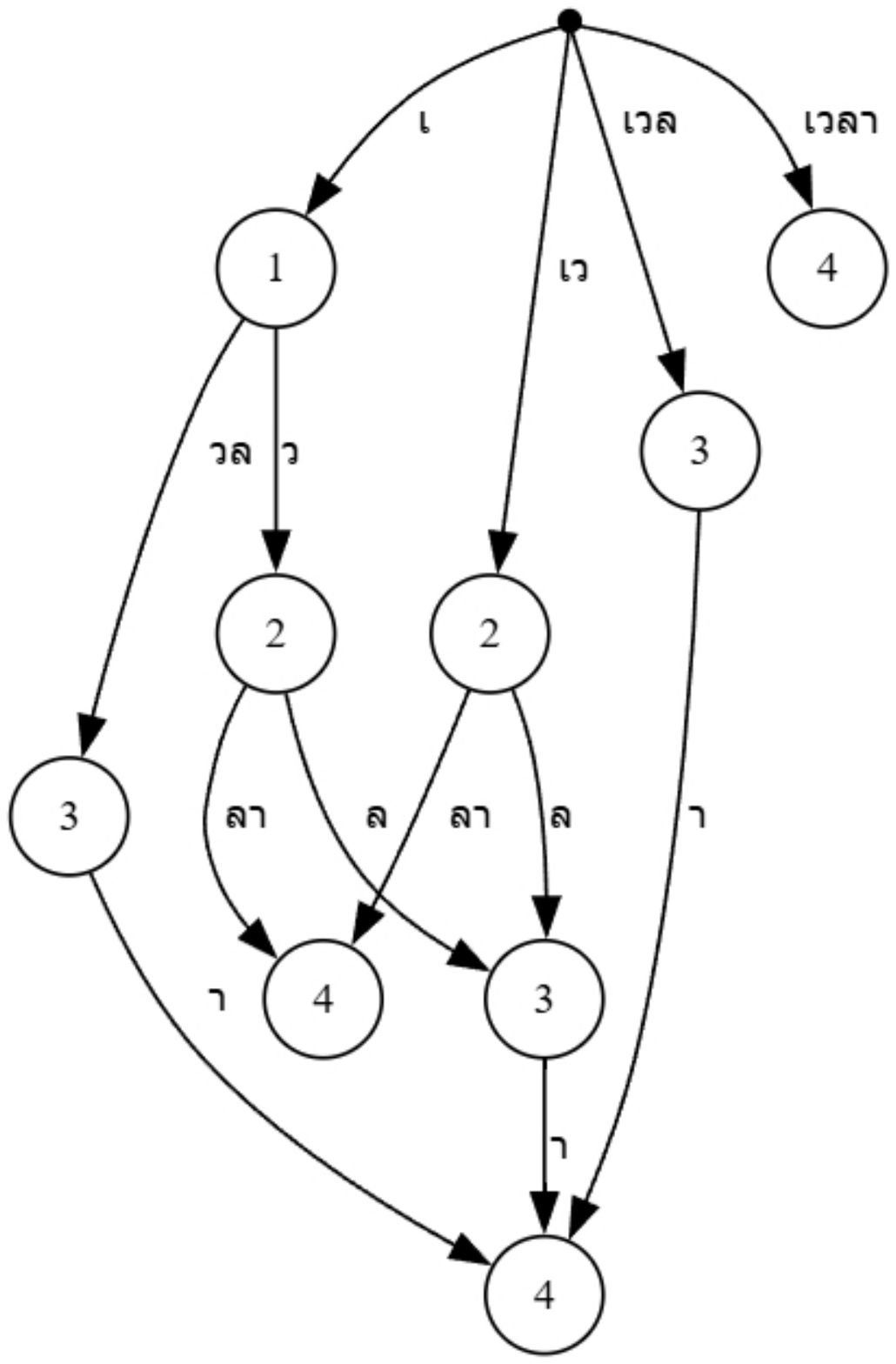}
    \caption{Thai}
\end{subfigure}
\hfill
\begin{subfigure}{0.23\textwidth}
    \centering
    \includegraphics[width=\linewidth]{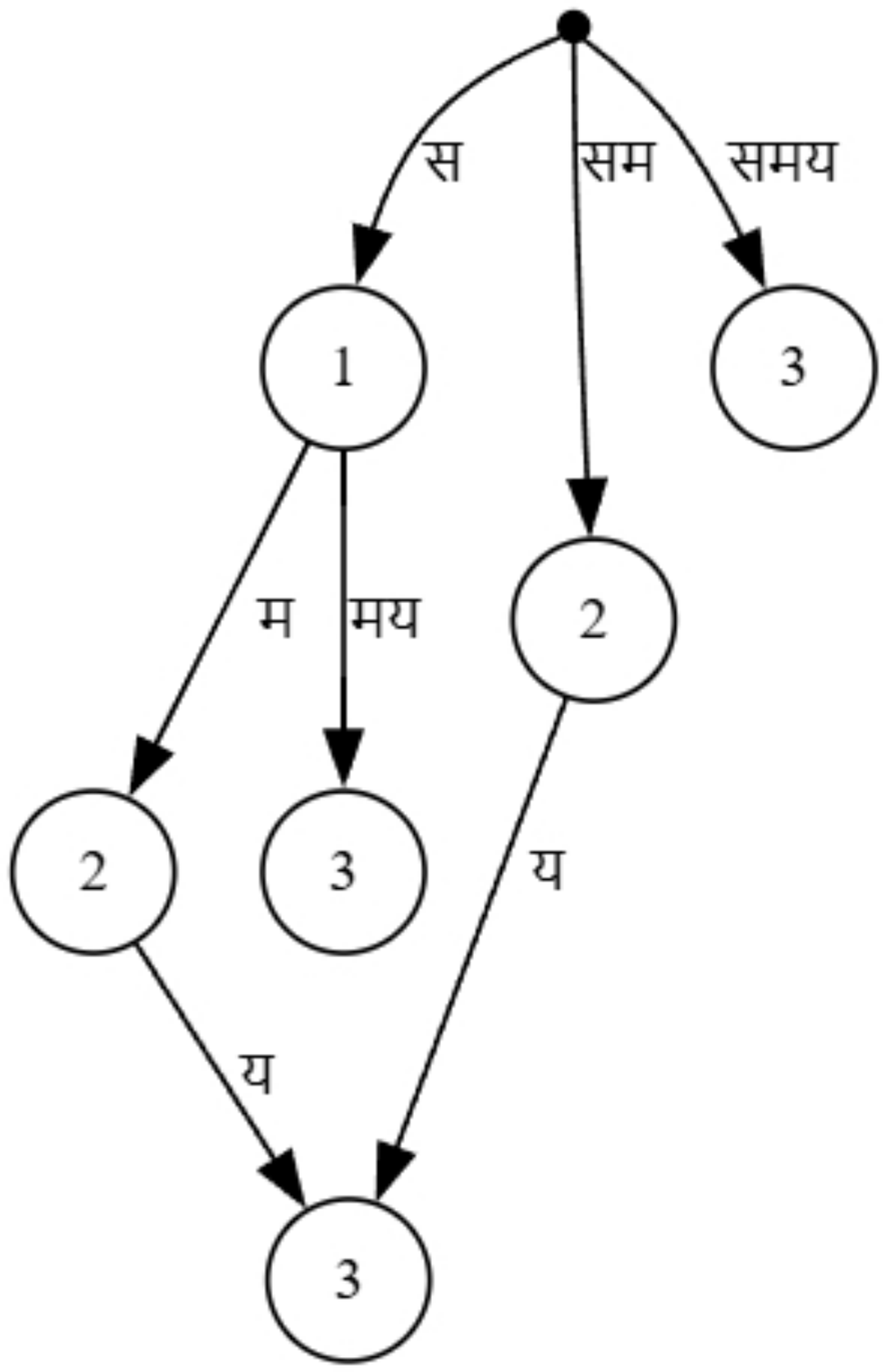}
    \caption{Hindi}
\end{subfigure}
\caption{MDD of the word ``time'' across different languages.}
\label{fig:mdds}
\end{figure*}
\paragraph{Tokenization types.}
We evaluate models under three tokenization settings.

\subparagraph{(1) Canonical tokenization.}
\[
t_c := T_{\mathrm{canon}}(x).
\]

This is the default segmentation used at inference time.

\subparagraph{(2) Random non-canonical tokenization.}
A non-canonical tokenization is any tokenization $t$ where
\[
t \in \mathcal{T}(x), \quad \text{s.t. } t \neq t_c
\]
Following \citet{geh2025adversarial}, 
we represent the space of all tokenizations using an Multi-valued Decision Diagram (MDD) as illustrated in Figure \ref{fig:mdds}. A Multi-valued Decision Diagram (MDD) can be used to represent all possible tokenizations of a given string.
In this representation, each node corresponds to a position in the input string, while edges between nodes are labeled with the tokens they represent. More specifically, an edge from node \(i\) to node \(j\) is associated with the token $x_{i:j} = (x_i, x_{i+1}, \ldots, x_j)$. 
Any path from the root node to a terminal node corresponds to one valid tokenization of the string \(x\). Using an MDD, we can uniformly sample from the space of all tokenizations by randomly selecting edges weighted by the number of possible tokenizations in the corresponding sub-MDD. We generate $N$ such samples:
\[
t_n \sim \mathcal{U}(\mathcal{T}(x)), n \in {1,\ldots,N}
\]
\noindent where $\mathcal{U}$ denotes sampling uniformly at random.
 

\subparagraph{(3) Character-level tokenization.}
Let $x = (x_1, \dots, x_{|x|})$. The character-level tokenization is
\[
t_{\mathrm{char}} = (x_1, x_2, \dots, x_{|x|}),
\]
where each symbol is treated as an individual token. By construction, $t_{\mathrm{char}} \in \mathcal{T}(x)$.

\paragraph{Evaluation.}
For a given string $x$, we compare model outputs across tokenizations:
\[
\mathrm{Err}(t_c, t_n), 
\quad 
\mathrm{Err}(t_c, t_{\mathrm{char}})
\]
All tokenizations are constrained to lie in $\mathcal{T}(x)$, ensuring the decoded string is identical and differences arise solely from token segmentation. Thus, any observed differences in 
$f(.)$ isolate the model’s sensitivity to tokenization.

\section{Experimental Setup}\label{experiment_setup}
\begin{figure}[t!]
\centering

\makebox[\columnwidth][c]{
\begin{subfigure}[c]{0.6\columnwidth}
    \centering
    \includegraphics[width=\linewidth]{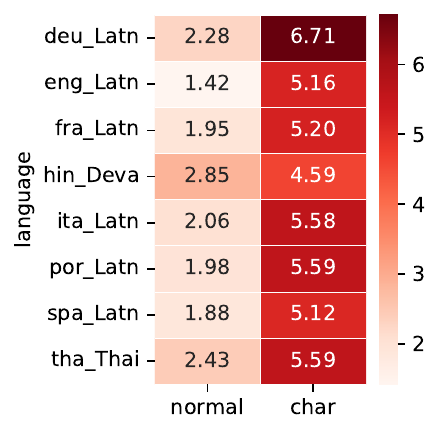}
    \caption{Llama}
    \label{fig:frag_llama}
\end{subfigure}
}

\vspace{0.5em}

\begin{subfigure}[c]{0.48\columnwidth}
    \centering
    \includegraphics[width=\linewidth]{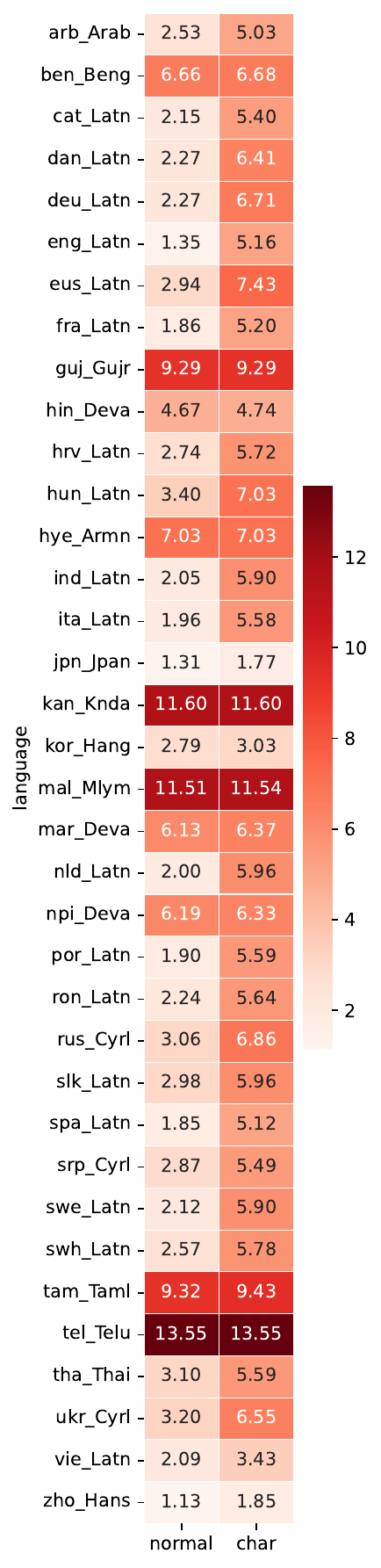}
    \caption{Qwen}
    \label{subfig:fragmentation_qwen}
\end{subfigure}
\hfill
\begin{subfigure}[c]{0.48\columnwidth}
    \centering
    \includegraphics[width=\linewidth]{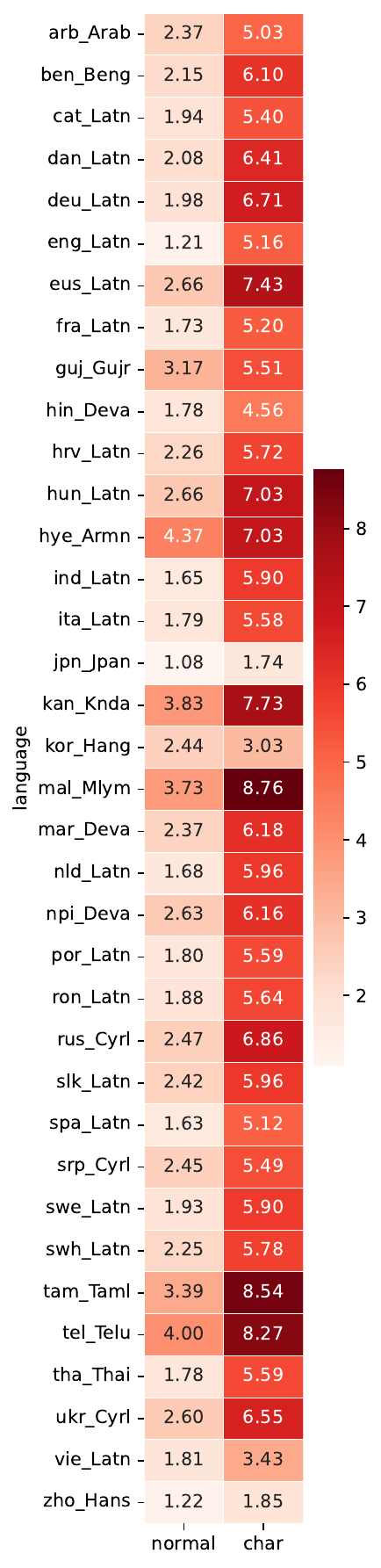}
    \caption{Gemma}
    \label{subfig:fragmentation_gemma}
\end{subfigure}

\caption{Fragmentation across languages under canonical and character-level tokenization, measured on the FLORES-200 benchmark.} 
\label{fig:fragmentation_heatmaps}
\end{figure}

We conduct experiments using the lm-evaluation-harness framework \cite{eval-harness}, which provides a standardized interface for evaluating language models across a wide range of benchmarks. We extend the framework to support alternative tokenization schemes by introducing preprocessing modules that generate non-canonical tokenizations and character-level tokenizations before model inference. These perturbations preserve the original text string but alter the token segmentation supplied to the model. For random non-canonical tokenization, we sample uniformly at random from the space of all tokenizations across 5 different seeds and report the mean.

\paragraph{Language Models.} The main experiments in this paper are conducted on several instruction-tuned models, namely Llama-3.1-8B-Instruct \cite{grattafiori2024llama}, Qwen3-8B \cite{yang2025qwen3}, and gemma-3-12b-it \cite{gemmateam2025gemma3technicalreport}, which are referred to throughout the paper as \textit{Llama}, \textit{Qwen}, and \textit{Gemma}, respectively.

\paragraph{Fragmentation Rate.} 
To quantify tokenization behavior, we compute the average number of tokens per word for each language. This metric measures the degree of token fragmentation produced by the tokenizer.

Before evaluating robustness to different tokenizations, we first analyze how the tokenizer segments text across languages under canonical and character-level tokenization using the many-to-many multilingual FLORES-200 benchmark \cite{costa2022no}, as illustrated in Figure \ref{fig:fragmentation_heatmaps}. Canonical tokenization represents the tokenizer’s default behavior, while character-level tokenization provides the maximum possible fragmentation. Together, they define the practical range of fragmentation exhibited by the tokenizer across languages.\footnote{In principle, tokenizations shorter than the canonical tokenization may exist; however, such cases are small in number, and hence we treat canonical and character-level tokenization as the practical lower and upper bounds of fragmentation.} 


\paragraph{Languages.} 
We conduct our main evaluation on languages that are supported across the models being compared to ensure fair cross-model comparisons. Since Gemma does not provide an official list of supported languages, we use the Qwen language set as a proxy. \footnote{Looking at the canonical performance of Gemma on multilingual ARC, we do see that many of the languages have reasonably good support (with >0.8 task performance).  Also for every other task in Table \ref{tab:main-results}, the performance on this subset of languages using Gemma is better than with using Qwen. This gives us confidence that these are indeed languages on which Gemma is reasonably proficient.}

As shown in Figure \ref{subfig:fragmentation_qwen}, several languages, namely Bengali, Gujarati, Armenian, Kannada, Malayalam, Marathi, Nepali, Tamil, and Telugu, exhibit canonical behavior that is nearly equivalent to character-level tokenization for Qwen. For these languages, canonical tokenization is already near the maximum fragmentation, leaving little room to construct meaningful non-canonical variants and making the comparison with other languages qualitatively different. Importantly, these languages are also not officially supported by Llama, leaving Gemma as the only model for which meaningful evaluation is possible. Since our goal is to compare robustness consistently across model families, we exclude these languages from the main analysis to ensure a fair and interpretable evaluation across languages. We present a separate analysis in Section \ref{sec:res_excluded_languages} in the Appendix to highlight their distinctive tokenization behavior. The final list of languages consists of 27 languages spanning diverse language families and scripts, as shown in Table \ref{tab:languages}.




\paragraph{Task.} 
We perform generation-based evaluation on both short-answer tasks (MLQA, MGSM) and multiple-choice tasks (Multi-ARC, Multi-HellaSwag \cite{lai-etal-2023-okapi}, Global-MMLU-Lite \cite{singh-etal-2025-global}, and Belebele \cite{bandarkar-etal-2024-belebele}). Details about tasks is present in Table \ref{tab:eval_tasks}.

We retokenize the entire prompt string, including the instruction and the multiple choice answer choices. We do not retokenize the system prompt or special tokens.

\begin{figure}[t!]
\centering

\begin{subfigure}{\linewidth}
    \centering
    \includegraphics[width=\linewidth]{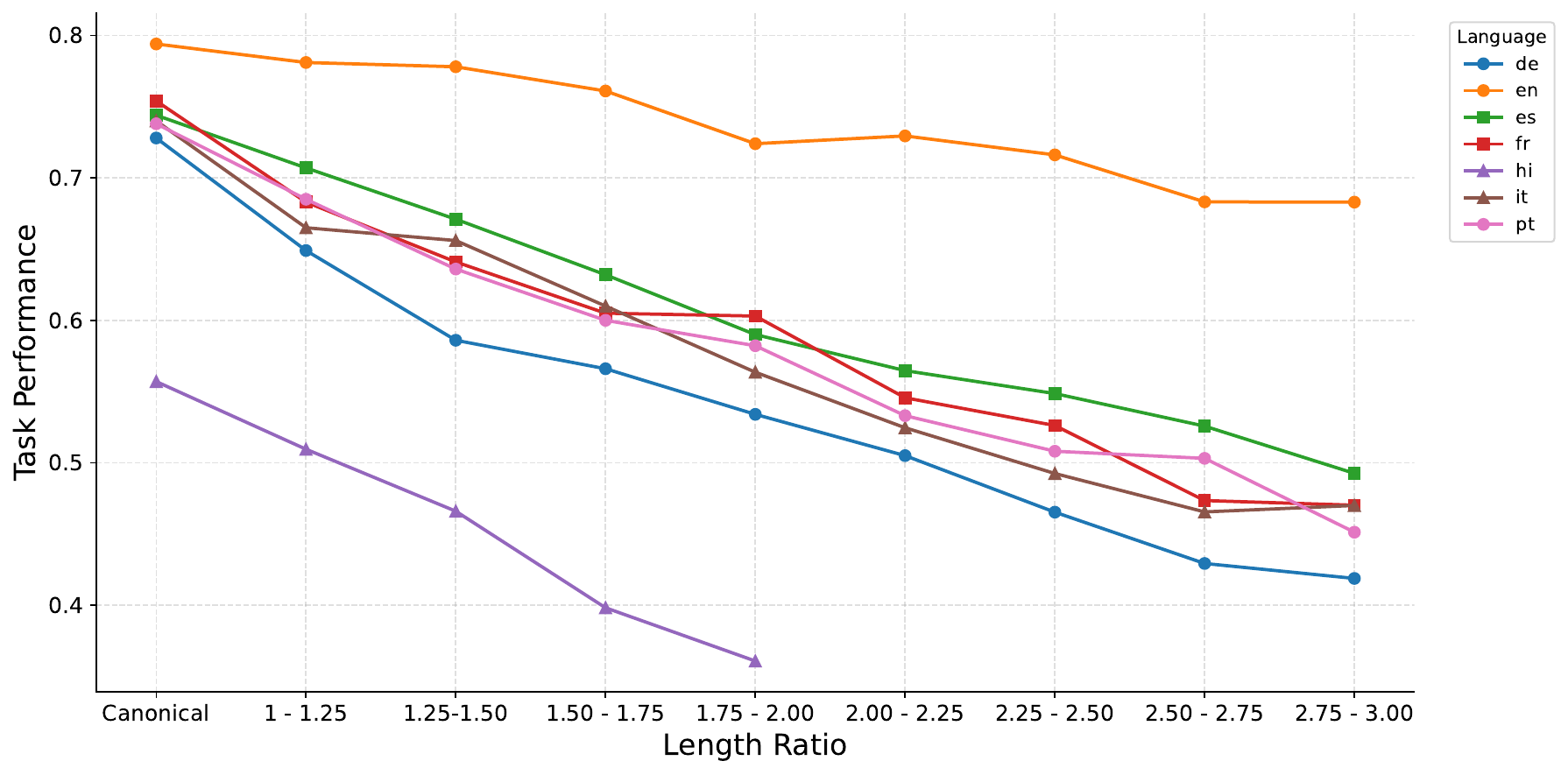}
    \caption{Llama}
\end{subfigure}

\vspace{0.5em}

\begin{subfigure}{\linewidth}
    \centering
    \includegraphics[width=\linewidth]{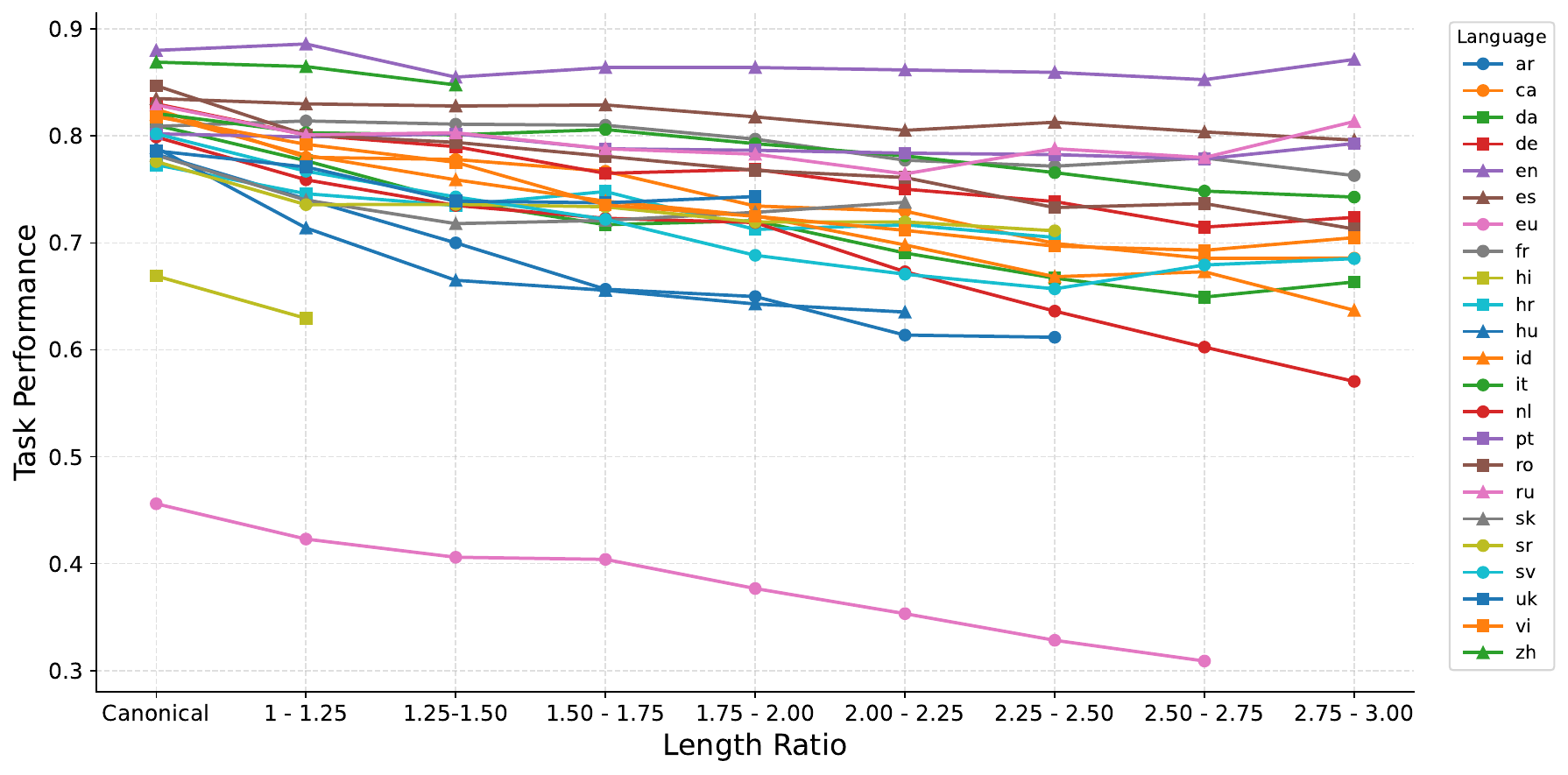}
    \caption{Qwen}
\end{subfigure}

\vspace{0.5em}

\begin{subfigure}{\linewidth}
    \centering
    \includegraphics[width=\linewidth]{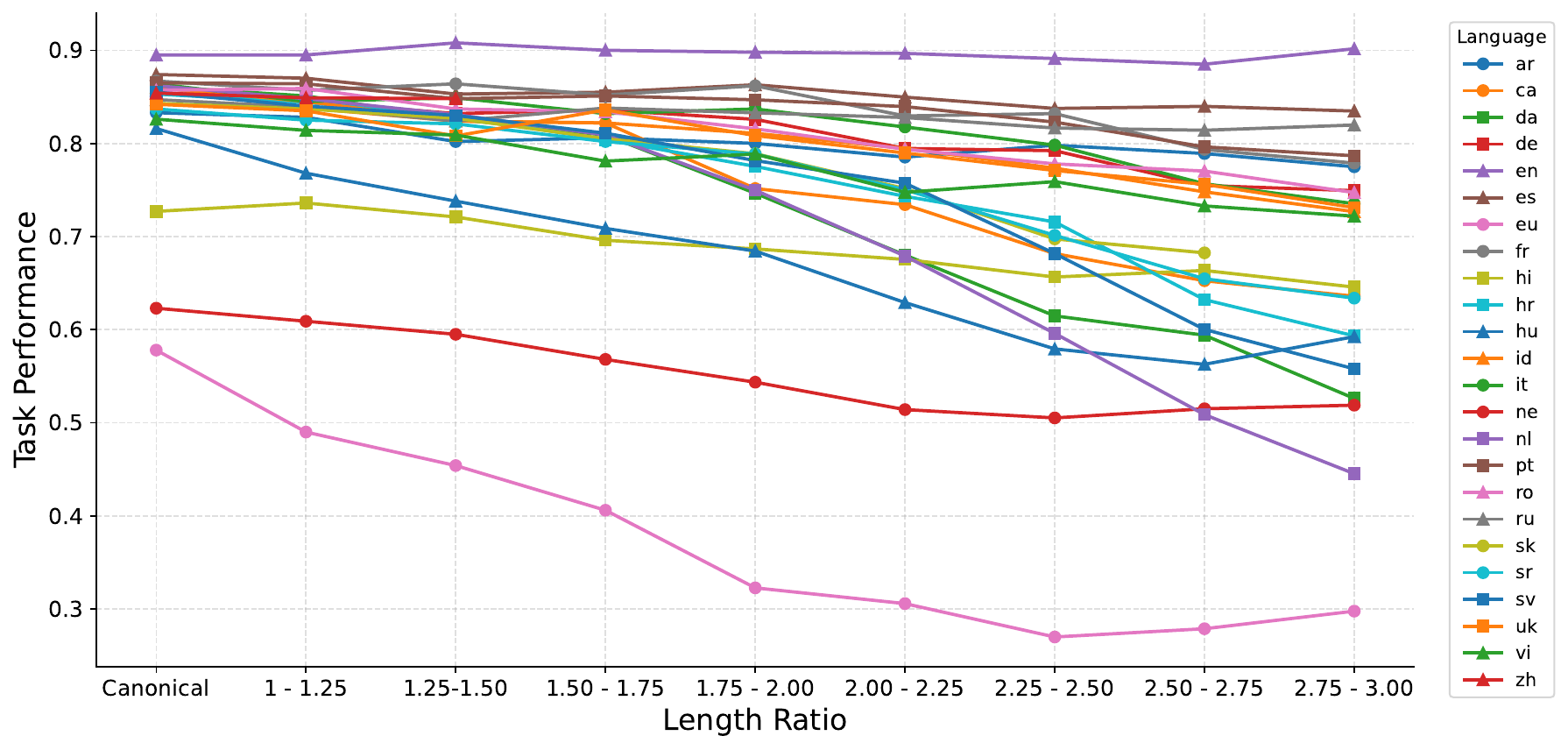}
    \caption{Gemma}
\end{subfigure}

\caption{Impact of granularity of non-canonical tokenization on performance on Multilingual ARC}
\label{fig:granularity_multi_arc}
\end{figure}

\section{Results}
\begin{table*}[t!]
\centering
\resizebox{\textwidth}{!}{%
\begin{tabular}{@{}c|ccc|ccc|ccc@{}}
\toprule
\multirow{2}{*}{\textbf{Benchmark}} &
  \multicolumn{3}{c|}{\textbf{Llama-3.1-8B-Instruct}} &
  \multicolumn{3}{c|}{\textbf{Qwen/Qwen3-8B}} &
  \multicolumn{3}{c}{\textbf{gemma-3-12b-it}} \\ \cmidrule(l){2-10} 
 &
  \textbf{Canon} &
  \textbf{Rand $\Delta$\%} &
  \textbf{Char $\Delta$\%} &
  \textbf{Canon} &
  \textbf{Rand $\Delta$\%} &
  \textbf{Char $\Delta$\%} &
  \textbf{Canon} &
  \textbf{Rand $\Delta$\%} &
  \textbf{Char $\Delta$\%} \\ \midrule
\textbf{MGSM}             & 62.64  & 27.522 & 48.661 & 45.911 & 19.948 & 21.856 & 68.622 & 7.89   & 17.609 \\
\textbf{MLQA}             & 42.474 & 24.446 & 41.424 & 58.05  & 12.002 & 13.58  & 70.075 & 14.708 & 15.675 \\ \midrule
\textbf{Global-MMLU-Lite} & 61.929 & 20.371 & 35.046 & 59.883 & 8.957  & 13.721 & 66.433 & 9.766  & 18.702 \\
\textbf{Multi-ARC}        & 72.214 & 24.32  & 41.232 & 78.275 & 8.542  & 15.963 & 82.729 & 9.865  & 26.726 \\
\textbf{Multi-HellaSwag}  & 64.486 & 25.754 & 43.81  & 65.17  & 12.202 & 21.071 & 70.978 & 10.012 & 27.258 \\
\textbf{Belebele}         & 79.764 & 20.221 & 35.586 & 81.201 & 6.692  & 10.353 & 87.412 & 7.382  & 19.597 \\ \midrule
\textbf{Average}          & 63.918 & 23.772 & 40.96  & 64.748 & 11.391 & 16.091 & 74.375 & 9.937  & 20.928 \\ \bottomrule
\end{tabular}%
}
\caption{Generation-based evaluation across multilingual benchmarks. Scores are averaged across languages. We report the relative performance drop from canonical tokenization under randomly sampled non-canonical tokenization (Rand $\Delta$\%) and character-level tokenization (Char $\Delta$\%). MLQA and MGSM are short-answer (SA) tasks; Multi-HellaSwag, Multi-ARC, Global-MMLU-Lite, and Belebele are multiple-choice (MC). Performance degrades when inputs use non-canonical tokenization.}
\label{tab:main-results}
\end{table*}

\subsection{Impact of non-canonical tokenization}
As observed in Table \ref{tab:main-results}, the results demonstrate that multilingual language models are highly sensitive to tokenization structure, even when the underlying input string remains semantically identical. Across all benchmarks, canonical tokenization consistently yields the highest performance, while both random non-canonical and character tokenizations lead to measurable degradation. Character tokenization causes substantially larger performance drops than random non-canonical tokenization for every model and task, indicating that increased segmentation fragmentation severely disrupts model representations. Among the evaluated models, Qwen3-8B exhibits the strongest robustness, with only 11.39\% average degradation under random tokenization and 16.09\% under character tokenization, suggesting stronger invariance to alternative token decompositions. In contrast, Llama-3.1-8B-Instruct shows the highest vulnerability, with degradation increasing from 23.77\% to 40.96\%, revealing a strong dependence on canonical subword segmentation. Gemma-3-12B-IT achieves the highest canonical performance overall but displays intermediate robustness, remaining stable under random tokenization while degrading more substantially under character-level fragmentation. At task-level, we observe that that short-answer benchmarks such as MGSM and MLQA are more sensitive to tokenization variation, whereas multi-choice benchmarks namely Global-MMLU, Multi-ARC, Multi-HellaSwag and Belebele remains comparatively robust across models.

For Multilingual-ARC, we analyze whether fragmentation rate is predictive of model robustness under non-canonical tokenization. For each example, we compute the canonical fragmentation rate (FR) and evaluate model performance under both canonical and non-canonical tokenizations. We then fit a logistic regression model where the dependent variable is non-canonical performance (correct/incorrect), and the predictors are canonical performance and canonical fragmentation rate.

For English, canonical performance is a strong positive predictor of non-canonical performance ($\beta = 3.42$, $p < 0.001$), indicating that examples answered correctly under canonical tokenization are substantially more likely to remain correct under non-canonical tokenizations. Canonical fragmentation rate exhibits a significant negative correlation with non-canonical performance ($\beta = -1.11$, $p = 0.021$), suggesting that examples with more fragmented canonical tokenizations are more vulnerable to performance degradation under non-canonical tokenizations. For analysis on more languages, please refer to Table~\ref{tab:multilingual_arc_logit}.


\subsection{Impact of granularity}
We study the impact of non-canonical tokenization granularity by sampling non-canonical tokenizations whose fragmentation ratio falls within a bucket relative to the canonical tokenization. Lower buckets correspond to tokenizations close to the canonical form, while higher buckets contain increasingly fragmented tokenizations approaching character-level segmentation. This helps analyze how increasing token fragmentation affects model performance and robustness. For each example, we compute the feasible range described in Section \ref{experiment_setup}, which identifies the set of bucket ranges from which a valid non-canonical tokenization can be sampled across different languages.
As shown in Figure \ref{fig:granularity_multi_arc}, the results 
reveal a strong and systematic relationship between tokenization granularity and model performance across all three language models. As tokenization moves progressively farther from the canonical segmentation towards increasingly fragmented non-canonical forms 
task performance generally declines across languages and models, demonstrating that multilingual performance is highly sensitive to token decomposition. Llama-3.1-8B-Instruct exhibits the steepest degradation overall, with performance dropping sharply as granularity increases. For example, languages such as German, French, Portuguese show substantial drop in performance as the granularity increases. 
English is comparatively more stable, 
suggesting that models are substantially more robust for high-resource languages that dominate pretraining corpora. 

Qwen3-8B demonstrates the strongest robustness and smoothest degradation curves across most languages. Several languages such as Portuguese, Russian, French, and English remain remarkably stable even at high granularity levels, with English preserving performance near canonical accuracy throughout the entire range. In contrast, lower-resource languages such as Basque, Dutch, and Hungarian show larger declines, indicating that robustness is correlated with language representation during training. Gemma-3-12B-IT achieves the highest canonical accuracies overall but exhibits mixed robustness patterns. High-resource languages including English, Spanish, German, French, and Russian remain relatively stable under moderate granularity increases, whereas languages such as Dutch, Swedish, Croatian, and Basque experience severe deterioration at higher bucket levels.

\paragraph{Controlling for Sequence Length} One possible explanation for the observed degradation under non-canonical tokenization is the increase in sequence length rather than changes in token boundaries themselves. To control for this effect, we repeat the Multilingual-ARC experiments using only non-canonical tokenizations whose lengths closely match the canonical tokenization (length-ratio buckets of 0.95--1 and 1--1.05). Results are present in Table \ref{tab:length_control} in the Appendix. Despite nearly identical sequence lengths, the performance gap largely persists across languages, indicating that the observed degradation cannot be explained solely by longer input sequences. These findings suggest that changes in token boundaries themselves contribute substantially to the loss in performance.


\begin{table*}[ht]
\centering
\resizebox{0.8\textwidth}{!}{%
\begin{tabular}{lcc cc cc}
\toprule
\multirow{2}{*}{Benchmark} &
\multicolumn{2}{c}{Qwen3-4B} &
\multicolumn{2}{c}{Qwen3-8B} &
\multicolumn{2}{c}{Qwen3-14B} \\
\cmidrule(lr){2-3}
\cmidrule(lr){4-5}
\cmidrule(lr){6-7}
& Canon & Rand $\Delta\%$
& Canon & Rand $\Delta\%$
& Canon & Rand $\Delta\%$ \\
\midrule
MGSM             & 44.267 & 21.085  & 45.911 & 19.948 & 45.289 & 16.208 \\
MLQA             & 47.564 & 14.648  & 58.050 & 12.002 & 65.329 & 10.841 \\
\midrule
Global-MMLU-Lite & 25.933 & -26.882 & 59.883 & 8.957  & 58.050 & 7.518 \\
Multi-ARC        & 38.255 & -27.978 & 78.275 & 8.542  & 78.400 & 7.169 \\
\bottomrule
\end{tabular}%
}
\caption{Results on scaling experiments. For multi-choice tasks, namely MMLU and ARC, Qwen3-4B model fails to follow system instruction and hence the results are arbitrary for small scale models. For short-answer tasks, we observe across tasks, that the robustness increases with scale.}
\label{tab:scaling-results}
\end{table*}

\subsection{Impact of scale}

As observed in Table \ref{tab:scaling-results}, the scaling experiments reveal several notable trends regarding the effect of model size on robustness. For multilingual question answering (\textbf{MLQA}), canonical performance consistently improves with scale, increasing from 47.56 for Qwen3-4B to 65.33 for Qwen3-14B. At the same time, the relative robustness degradation under randomization gradually decreases from 14.65\% to 10.84\%, indicating that larger models are not only more accurate but also modestly more robust to non-canonically tokenized context.

A similar trend is observed on \textbf{MGSM}, where canonical performance remains nearly constant across model scales (44--45), while Rand $\Delta\%$ steadily decreases from 21.09\% for the 4B model to 16.21\% for the 14B model. This suggests that scaling primarily improves robustness on multilingual mathematical reasoning rather than baseline task performance.

For multiple-choice reasoning benchmarks such as \textbf{Global-MMLU-Lite} and \textbf{Multi-ARC}, the 4B model exhibits unstable and unreliable behavior, often failing to follow system instructions, which results in negative or highly arbitrary robustness scores. In contrast, the 8B and 14B models show more consistent behavior and achieve substantially higher canonical performance while maintaining low robustness degradation. This indicates that scaling primarily improves instruction-following and output consistency for smaller models rather than robustness itself.

Overall, the results provide evidence that increasing model scale improves both baseline task performance and robustness. The most significant gains occur when scaling from 4B to 8B parameters, whereas the improvements from 8B to 14B are comparatively modest, suggesting diminishing returns with further increases in model size.

\begin{figure}[t!]
\centering

\begin{subfigure}{0.99\columnwidth}
    \centering
    \includegraphics[width=\linewidth]{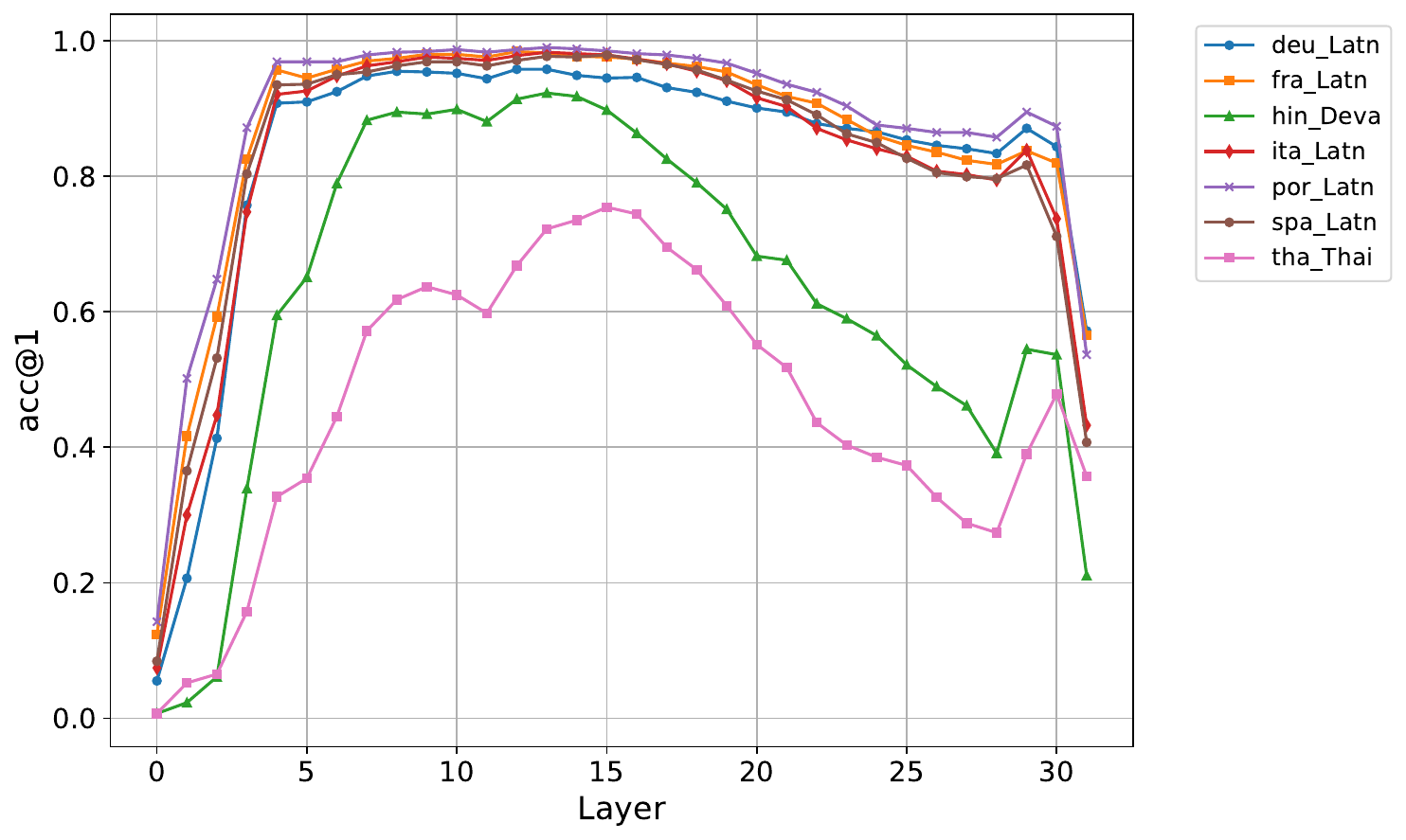}
    \caption{Normal}
\end{subfigure}

\vspace{0.5em}

\begin{subfigure}{0.99\columnwidth}
    \centering
    \includegraphics[width=\linewidth]{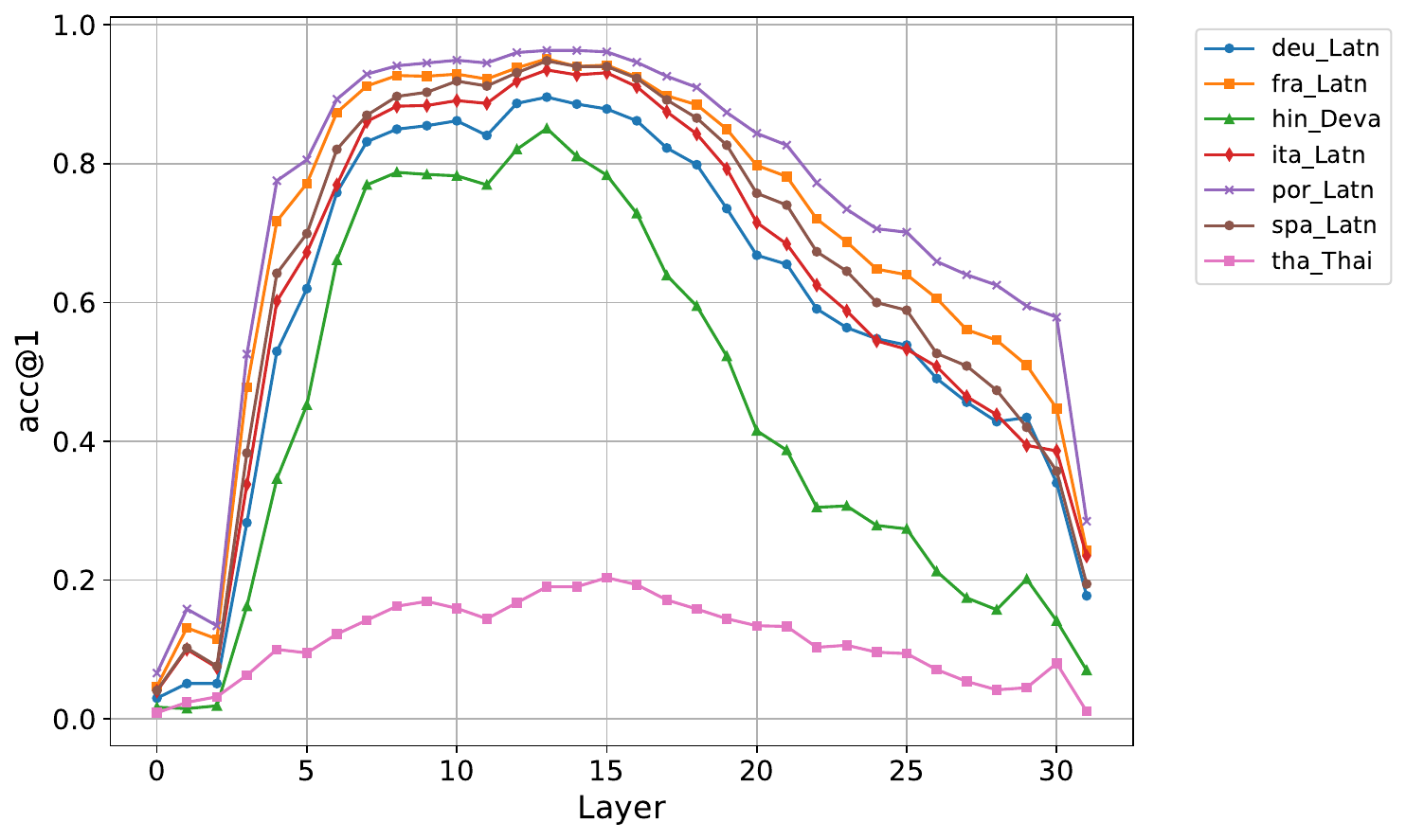}
    \caption{Random}
\end{subfigure}

\vspace{0.5em}

\begin{subfigure}{0.99\columnwidth}
    \centering
    \includegraphics[width=\linewidth]{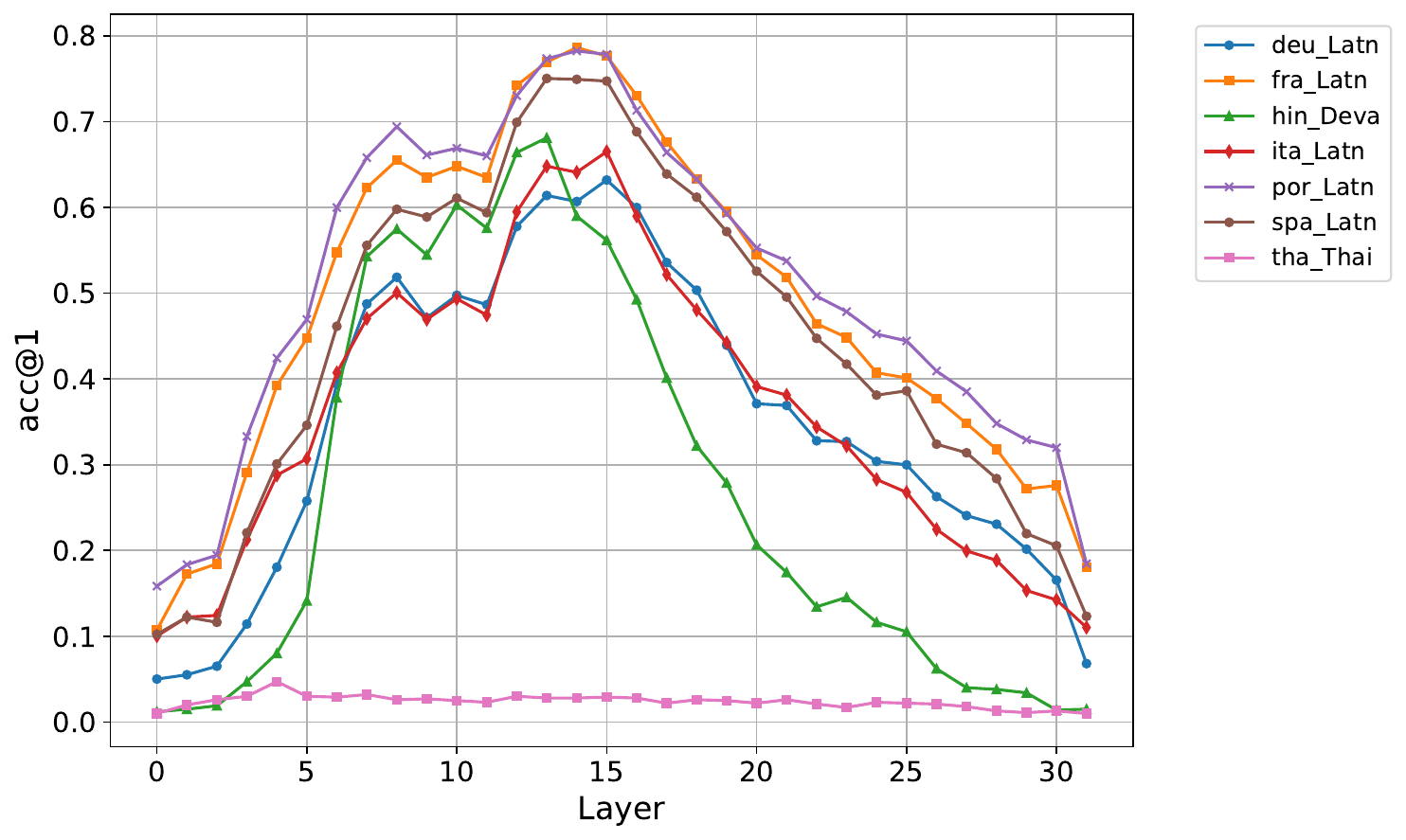}
    \caption{Character}
\end{subfigure}

\caption{Retrieval experiments on FLORES-200 for Llama}
\label{fig:retrieval-llama}
\end{figure}

\subsection{Cross-lingual experiments}
In addition to benchmark evaluation, we further probe into how token segmentations influence the LLM's internal representations. We perform a cross-lingual sentence retrieval experiment using the FLORES-200 multilingual benchmark dataset.
Following the standard retrieval setup 
we construct bilingual sentence pairs and evaluate retrieval accuracy. Given a query sentence in a source language, we compute its embedding and retrieve the closest sentence in the target language using cosine similarity, 
\[
\text{sim}(\mathbf{x}, \mathbf{y}) =
\frac{\mathbf{x} \cdot \mathbf{y}}
{\|\mathbf{x}\| \, \|\mathbf{y}\|}
\]
where $\mathbf{x}$ and $\mathbf{y}$ denote the sentence embeddings of the source sentence $x$ and target sentence $y$, respectively. 
We report top-1 retrieval accuracy, which measures the proportion of queries for which the correct translation is ranked highest among the candidates. To evaluate robustness to tokenization perturbations, we compute sentence embeddings under different tokenization settings (canonical, non-canonical, and character-level). Because the underlying text remains unchanged across conditions, differences in retrieval performance reflect changes in how token segmentation influences the model’s multilingual representations. Sentence embeddings are obtained by the final hidden state of the last token of the model. 

The results for Llama, Qwen, and Gemma are presented in Figures \ref{fig:retrieval-llama}, \ref{fig:retrieval-qwen}, and \ref{fig:retrieval-gemma}, respectively. Across all three models, canonical tokenization consistently gives the highest cross-lingual retrieval accuracy, while random and character-level tokenization reduce performance. High-resource European languages achieve the best retrieval scores but also suffer the largest degradation under tokenization perturbations, showing strong dependence on subword segmentation. In contrast, character-based languages such as Chinese, Japanese, and Korean are generally more robust, with smaller drops under character tokenization. Thai is the most sensitive language across models, while low-resource languages such as Yoruba remain difficult regardless of tokenization setting.

Among the models, Qwen3-8B is the most robust to tokenization changes, maintaining relatively strong retrieval performance even under character tokenization. Llama-3.1-8B-Instruct achieves high peak retrieval accuracy under canonical tokenization but shows degradation under non-canonical tokenization. Gemma-3-12B-it performs worst overall and is especially sensitive to tokenization. Overall, the experiments show that multilingual retrieval quality depends not only on semantics but also heavily on tokenizer robustness across scripts and language families.

\section{Non-Canonical Tokenization as a Source of Data Augmentation} 
In this section, our goal is to demonstrate a practical use of the exponentially large non-canonical tokenization space. We show that exposing the model to multiple tokenization variants of the same training examples acts as an effective form of data augmentation, improving multilingual performance and robustness through cross-lingual transfer. Non-canonical tokenization serves as a natural source of data augmentation for improving multilingual generalization.

\subsection{Tokenization Schemes}\label{sec:tokenization_schemes}
We investigate a family of tokenization schemes for constructing the training datasets used for LoRA fine-tuning. Each scheme specifies how tokenized representations are generated for every sentence in the original training corpus. For the multi-copy schemes ($N\times$, where $N \in \{10,20,30\}$), each original sentence is represented by $N$ tokenized variants, resulting in a proportionally larger training dataset.

The tokenization schemes are defined as follows:

\begin{itemize}
    \item \textbf{Canonical (1$\times$):} Each sentence is represented once using the tokenizer's canonical tokenization.

    \item \textbf{Non-Canonical (1$\times$):} Each sentence is represented once using a randomly sampled valid non-canonical tokenization.

    \item \textbf{Character (1$\times$):} Each sentence is represented once using the character-level tokenization, corresponding to the longest valid tokenization.

    \item \textbf{Canonical ($N\times$):} Each sentence is represented $N$ times, with every copy using the canonical tokenization.

    \item \textbf{Mixed ($N\times$):} Each sentence is represented using repeated groups of ten tokenized variants. Each group consists of one canonical tokenization, one character-level tokenization, and eight independently sampled non-canonical tokenizations. For larger values of $N$, this ten-copy pattern is repeated (e.g., twice for $20\times$, three times for $30\times$, and so on).

    \item \textbf{LR-Bucketed ($N\times$):} Each sentence is represented using repeated groups of ten tokenized variants. Each group consists of one canonical tokenization, one character-level tokenization, and eight non-canonical tokenizations sampled from eight disjoint token-length-ratio buckets. The token-length ratio is defined as the ratio of the number of tokens in the sampled tokenization to that of the canonical tokenization. The eight buckets correspond to the intervals $[1.00,1.25)$, $[1.25,1.50)$, $[1.50,1.75)$, $[1.75,2.00)$, $[2.00,2.25)$, $[2.25,2.50)$, $[2.50,2.75)$, and $[2.75,3.00]$. For larger values of $N$, this ten-copy pattern is repeated accordingly.
\end{itemize}

\subsection{Dataset}
Among all the datasets considered in our study, \textbf{Multilingual-ARC} is the only dataset that provides an official training split. Consequently, all LoRA fine-tuning experiments are conducted exclusively on Multilingual-ARC. The original training split consists of 1,116 training instances and 298 validation instances. We apply the tokenization schemes described in Section~\ref{sec:tokenization_schemes} to construct the corresponding training datasets for fine-tuning.\\
For model selection, we construct a validation set in which each original validation instance is represented by its canonical, sampled non-canonical, and character tokenizations, yielding a total of 894 validation instances. Validation is performed on this unified validation set for all training schemes.Details about hyperparamters are present in Table \ref{tab:lora-hyperparameters}.

\begin{table}[]
\centering
\resizebox{\columnwidth}{!}{%
\begin{tabular}{@{}c|ccc|cc@{}}
\toprule
\textbf{Training Scheme}    & \textbf{Canon} & \textbf{Rand} & \textbf{Char} & \textbf{Rand $\Delta$\%} & \textbf{Char $\Delta$\%} \\ \midrule
\textbf{Baseline (No Fine-Tuning)} & 78.3 & 71.8 & 66   & 8.542 & 15.963 \\ \midrule
\textbf{Canonical (1×)}            & 79.2 & 72.3 & 66.3 & 8.97  & 16.502 \\
\textbf{Non-Canonical (1×)}        & 78.8 & 72.1 & 66.3 & 8.713 & 16.129 \\
\textbf{Character (1×)}            & 78.8 & 72   & 66.2 & 8.932 & 16.271 \\ \midrule
\textbf{Canonical (10 ×)}          & 81.1 & 73.6 & 67.7 & 9.436 & 16.524 \\
\textbf{Mixed (10 ×)}              & 81.1 & 73.7 & 67.9 & 9.2   & 16.28  \\
\textbf{LR-Bucketed (10 ×)}        & 81.1 & 73.8 & 68   & 9.131 & 16.151 \\ \midrule
\textbf{Canonical (20 ×)}          & 81.2 & 73.7 & 67.9 & 9.34  & 16.399 \\
\textbf{Mixed (20 ×)}              & 81.5 & 74.6 & 68.7 & 8.66  & 15.783 \\
\textbf{LR-Bucketed (20 ×)}        & 81.6 & 74.6 & 68.9 & 8.78  & 15.701 \\ \midrule
\textbf{Canonical (30 ×)}          & 81.2 & 73.7 & 67.9 & 9.405 & 16.49  \\
\textbf{Mixed (30 ×)}              & 81.2 & 74   & 68.2 & 8.987 & 16.141 \\
\textbf{LR-Bucketed (30 ×)} & \textbf{81.8}  & \textbf{74.8} & \textbf{69.2} & 8.637             & 15.548            \\ \bottomrule
\end{tabular}%
}
\caption{Average accuracy (\%) across all languages on the Multilingual-ARC test set under canonical, randomly sampled non-canonical, and character-level tokenizations. \textit{Rand} $\Delta$ and \textit{Char} $\Delta$ denote the average relative percentage decrease in accuracy with respect to canonical evaluation for the corresponding training scheme. Impact across languages can be observed in Figure \ref{fig:lora_languages} in the Appendix.}
\label{tab:lora-results}
\end{table}

\section{Analysis}
As observed in Table \ref{tab:lora-results}, the results demonstrate that the proposed multi-tokenization training schemes consistently improve cross-lingual generalization and robustness to tokenization variation. Although LoRA fine-tuning is performed exclusively on the English training split of Multilingual-ARC, performance improves consistently across all languages across all evaluation settings. 
This indicates that the benefits of multi-tokenization training transfer effectively across languages and are not confined to the language observed during fine-tuning.

We can observe that the choice of tokenization has only a limited effect when each training instance is represented by a single tokenization. The Canonical (1$\times$), Non-Canonical (1$\times$), and Character (1$\times$) schemes achieve nearly identical average performance across all three evaluation settings, suggesting that simply replacing one tokenization with another provides little additional robustness.

In contrast, increasing the number of tokenized representations per training instance yields substantial improvements. Relative to Canonical (1$\times$), Canonical (10$\times$) improves the average accuracy from 0.792 to 0.811 under Canonical evaluation, from 0.723 to 0.736 under Non-Canonical evaluation, and from 0.663 to 0.677 under Character evaluation. These gains demonstrate that exposing the model to multiple valid tokenizations of the same training instance is considerably more effective. 

Increasing the number of tokenized variants from 10$\times$ to 20$\times$ produces additional but smaller improvements, indicating diminishing returns with increasing dataset expansion. For example, the Mixed scheme improves from $(0.811,0.737,0.679)$ to $(0.815,0.746,0.687)$ under Canonical, Non-Canonical, and Character evaluation, respectively, while LR-Bucketed improves from $(0.811,0.738,0.680)$ to $(0.816,0.746,0.689)$. These improvements are modest compared to those obtained when moving from 1$\times$ to 10$\times$, suggesting that further dataset expansion provides only incremental improvements. 

It is important to note the comparison between the Mixed and LR-Bucketed schemes highlights the importance of the distribution of non-canonical tokenizations presented during training. While both schemes consistently outperform Canonical training at the same dataset multiplier, LR-Bucketed achieves the highest average performance under canonical (0.816) and character (0.689) evaluation, while matching mixed under non-Canonical evaluation (0.746). These results suggest that systematically covering the space of valid non-canonical tokenizations through token-length-ratio buckets is more effective than repeatedly sampling arbitrary non-canonical tokenizations. Moreover, as observed in Table \ref{tab:lora-results}, although Canonical scheme improves performance, it comes at the expense of robustness. In contrast, LR-Bucketed schemes achieve even greater performance gains while preserving robustness. Taken together, the results indicate that robustness to tokenization variation is driven not only by increasing the number of tokenized training instances but also by increasing the diversity and coverage of the non-canonical tokenization space. 

\section{Conclusion}
In conclusion, although a given string can admit exponentially many valid tokenizations, language models process only a single deterministic canonical tokenization, leaving the broader tokenization space insufficiently explored. This work systematically examines model behavior under non-canonical tokenizations across 27 languages and six downstream tasks, revealing that tokenization invariance observed in English does not generalize universally. Instead, language models exhibit substantial performance degradation under alternative tokenizations.Our results demonstrate that LoRA fine-tuning with multi-tokenization training is an effective strategy for mitigating tokenization sensitivity. Although fine-tuning is performed solely on English data, it consistently improves robustness across languages, with systematically sampling diverse non-canonical tokenizations yielding the best overall performance.

\section*{Limitations}
Our study has two main limitations. First, we evaluate non-canonical tokenization as a training-time data augmentation strategy only on Multilingual-ARC, as it is the only benchmark among those considered in this work that provides a training split. Evaluating this approach across a broader range of multilingual tasks would help establish the generality of the observed gains.

Second, although our analysis reveals systematic differences in robustness across models and languages, it does not isolate the individual factors responsible for these differences. Robustness may be influenced by multiple interacting factors, including vocabulary allocation, language resource level, pretraining representation etc. Disentangling the independent contribution of these factors would require carefully controlled experiments that manipulate each variable in isolation. We leave such analyses to future work.




\section*{Ethical Considerations}
Non-canonical tokenizations can affect model behavior and may increase the likelihood of incorrect, hallucinated, or harmful outputs. Our work investigates the robustness of language models to alternative valid tokenizations in order to better understand their sensitivity to input segmentation across languages. We do not intend these perturbation methods for harmful use; rather, we hope they help identify model vulnerabilities and motivate the development of more robust and reliable multilingual language models.



\bibliography{custom}

\appendix

\section{Model Size And Budget}
All experiments were conducted on NVIDIA RTX 6000 Ada Generation GPUs with 48 GB VRAM.

\section{Use of Large Language Models}
We used large language models (LLMs) to improve the grammar, clarity, and readability of the paper. All generated content was carefully reviewed and verified by the authors to ensure accuracy and consistency with the scientific contributions of the work.

\section{Results on Excluded Languages}\label{sec:res_excluded_languages}
As shown in Table~\ref{tab:excluded_tokenization_space}, the excluded languages have a substantially smaller tokenization space under the Qwen tokenizer because their canonical tokenization is already close to maximum fragmentation. Consequently, non-canonical tokenizations introduce only limited additional perturbation, which is reflected in the Multilingual-ARC results (Table~\ref{tab:excluded_languages_arc}), where Qwen3-8B exhibits only minor differences between canonical, non-canonical, and character-level tokenizations. In contrast, Gemma3-12B admits a much larger tokenization space for these languages, leading to greater variation in performance across tokenization schemes.

For Armenian, many sentences in the dataset have only a single valid tokenization under the Qwen3 tokenizer. Consequently, results for Armenian are omitted from these tables.

\begin{table}[t]
\centering
\small
\begin{tabular}{l|ccc|ccc}
\toprule
& \multicolumn{3}{c|}{\textbf{Qwen3-8B}} & \multicolumn{3}{c}{\textbf{Gemma3-12B}} \\
\cmidrule(lr){2-4} \cmidrule(l){5-7}
\textbf{Lang} & \textbf{Canon} & \textbf{Rand} & \textbf{Char} & \textbf{Canon} & \textbf{Rand} & \textbf{Char} \\
\midrule
bn & 57.70 & 52.64 & 53.00 & 65.61 & 53.80 & 46.96 \\
gu & 54.30 & 52.64 & 52.80 & 60.78 & 56.62 & 50.34 \\
kn & 47.40 & 44.08 & 43.80 & 56.71 & 44.64 & 39.55 \\
ml & 38.80 & 33.14 & 34.00 & 46.23 & 40.12 & 35.46 \\
ne & 57.40 & 49.84 & 48.10 & 62.53 & 55.02 & 47.05 \\
ta & 36.10 & 31.40 & 31.80 & 40.81 & 33.66 & 31.26 \\
te & 42.70 & 41.00 & 42.00 & 44.82 & 40.32 & 35.96 \\
\bottomrule
\end{tabular}
\caption{Results on Multilingual ARC for languages excluded from the main analysis.}
\label{tab:excluded_languages_arc}
\end{table}

\begin{table*}[t]
\centering
\small
\resizebox{0.45\textwidth}{!}{%
\begin{tabular}{l|cc}
\toprule
\textbf{Language} & \textbf{Qwen3-8B} & \textbf{Gemma3-12B} \\
\midrule
bn & 3.53  & 64.45 \\
gu & 1.12  & 59.73 \\
kn & 9.16  & 54.32 \\
ml & 6.33  & 64.18 \\
ne & 15.73 & 61.95 \\
ta & 10.08 & 71.92 \\
te & 17.07 & 56.19 \\
\bottomrule
\end{tabular}%
}
\caption{Average number of valid tokenizations ($\log_{10}$ scale) for the excluded languages under the Qwen3-8B and Gemma3-12B tokenizers measured on the FLORES-200 benchmark. The substantially smaller tokenization space for Qwen reflects its near character-level canonical tokenization for these languages.}
\label{tab:excluded_tokenization_space}
\end{table*}

\begin{table*}[t]
\centering
\scriptsize
\resizebox{0.45\textwidth}{!}{%
\begin{tabular}{lccc}
\toprule
\textbf{Lang} & \textbf{Canonical} & \textbf{0.95--1.00} & \textbf{1.00--1.05} \\
\midrule
ar & 78.30 & 74.73 & 73.00 \\
ca & 82.50 & -- & 80.10 \\
da & 81.00 & -- & 78.18 \\
de & 83.00 & 76.72 & 81.50 \\
en & 88.00 & -- & 87.90 \\
es & 83.50 & -- & 82.00 \\
eu & 45.60 & 40.75 & 41.10 \\
fr & 80.90 & -- & 81.30 \\
hi & 66.90 & -- & 62.51 \\
hr & 77.30 & 67.83 & 74.20 \\
hu & 78.90 & 73.33 & 71.10 \\
id & 82.00 & -- & 78.00 \\
it & 82.10 & 76.92 & 81.20 \\
ne & 57.60 & -- & 50.79 \\
nl & 79.90 & -- & 76.80 \\
pt & 80.20 & -- & 78.80 \\
ro & 84.70 & -- & 81.80 \\
ru & 82.90 & 77.64 & 81.10 \\
sk & 78.20 & 73.68 & 75.10 \\
sr & 77.60 & 81.25 & 74.98 \\
sv & 80.20 & 80.20 & 77.60 \\
uk & 78.60 & 70.87 & 75.80 \\
vi & 81.80 & -- & 79.42 \\
zh & 86.90 & 87.38 & 86.69 \\
\bottomrule
\end{tabular}%
}
\caption{Multilingual-ARC performance under canonical tokenization and non-canonical tokenizations with sequence lengths closely matching the canonical tokenization (length-ratio buckets of 0.95--1.00 and 1.00--1.05). Performance differences largely persist despite nearly identical sequence lengths.}
\label{tab:length_control}
\end{table*}

\section{Hyperparameter Details}
The hyperparameter settings for the LoRA fine-tuning experiments are provided in Table \ref{tab:lora-hyperparameters}.

\begin{table*}[h]
\centering
\resizebox{0.45\textwidth}{!}{%
\begin{tabular}{ll}
\toprule
\textbf{Hyperparameter} & \textbf{Value} \\
\midrule
Maximum sequence length & 2048 \\
LoRA rank ($r$) & 16 \\
LoRA scaling factor ($\alpha$) & 32 \\
LoRA dropout & 0.1 \\
LoRA target modules & $q_{\text{proj}}$, $k_{\text{proj}}$, $v_{\text{proj}}$ \\
Learning rate & $2.5 \times 10^{-6}$ \\
Number of epochs & 5 \\
Per-device batch size & 4 \\
Gradient accumulation steps & 16 \\
Evaluation interval & Every 25 steps \\
\bottomrule
\end{tabular}%
}
\caption{LoRA fine-tuning hyperparameters used in all experiments.}
\label{tab:lora-hyperparameters}
\end{table*}

\begin{table*}[t]
\centering
\small
\begin{tabular}{p{2.5cm} p{3cm} p{2cm} p{8cm}}
\hline
\textbf{Type} & \textbf{Task} & \textbf{No. of samples} & \textbf{System prompt} \\
\hline

Short answer 
& MLQA 
& 1000 
& You are a helpful QA assistant. Extract the answer span from the context. \\

& MGSM 
& 250 
& You are a helpful assistant. \\

\hline

Multiple choice 
& Multi-ARC 
& 1000 
& \multirow{4}{=}{You are a helpful assistant. For each of the following multiple-choice questions, respond with the correct answer letter (A, B, C, or D) only. Do not include any explanation, reasoning, or additional text.} \\

& Multi-HellaSwag 
& 1000 
& \\

& Global-MMLU-Lite 
& 400 
& \\

& Belebele 
& 900 
& \\

\hline
\end{tabular}
\caption{Tasks}
\label{tab:eval_tasks}
\end{table*}

\begin{figure*}[t!]
\centering

\begin{subfigure}{0.9\linewidth}
    \centering
    \includegraphics[width=\linewidth]{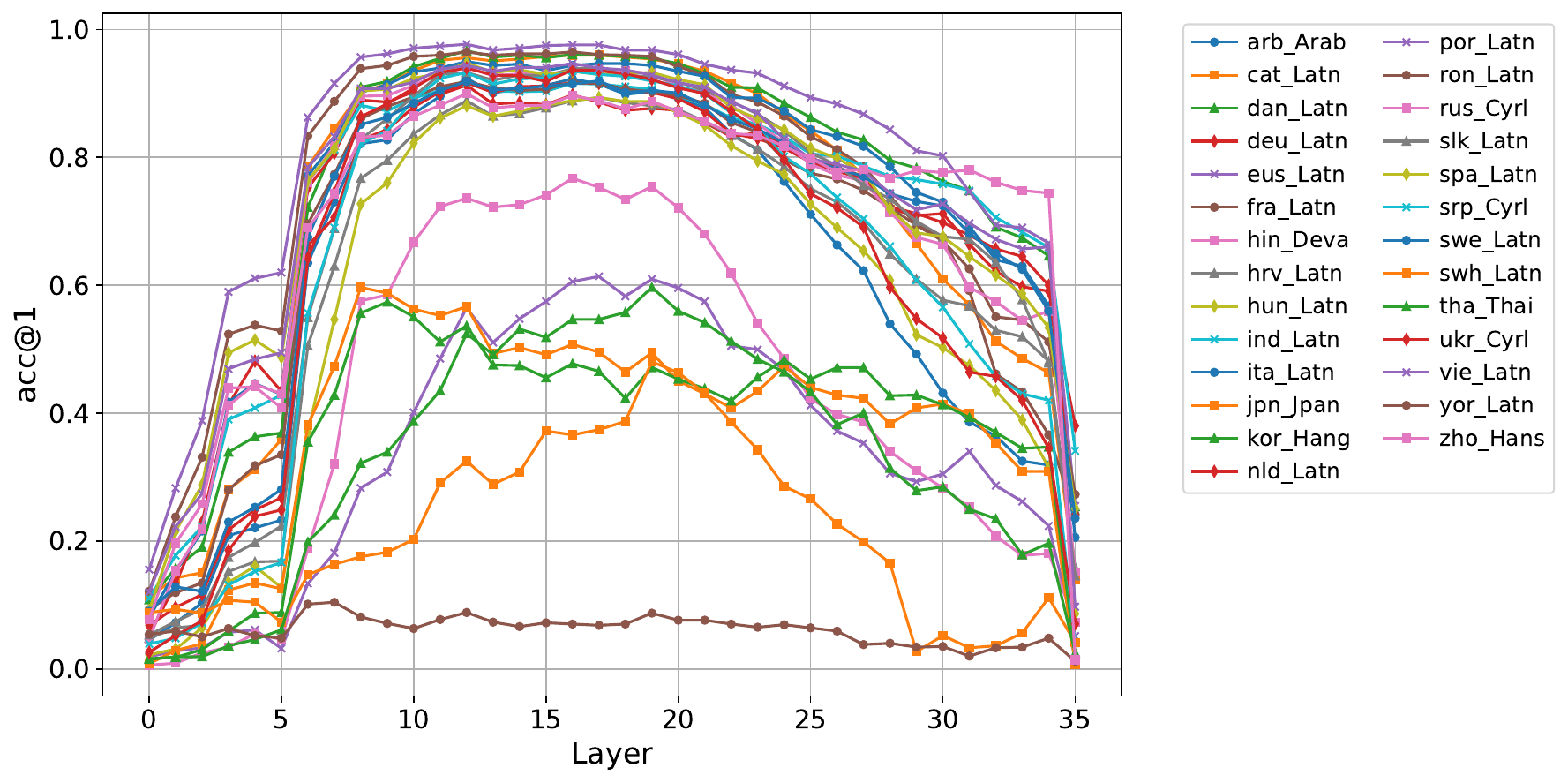}
    \caption{Normal}
\end{subfigure}

\vspace{0.5em}

\begin{subfigure}{0.9\linewidth}
    \centering
    \includegraphics[width=\linewidth]{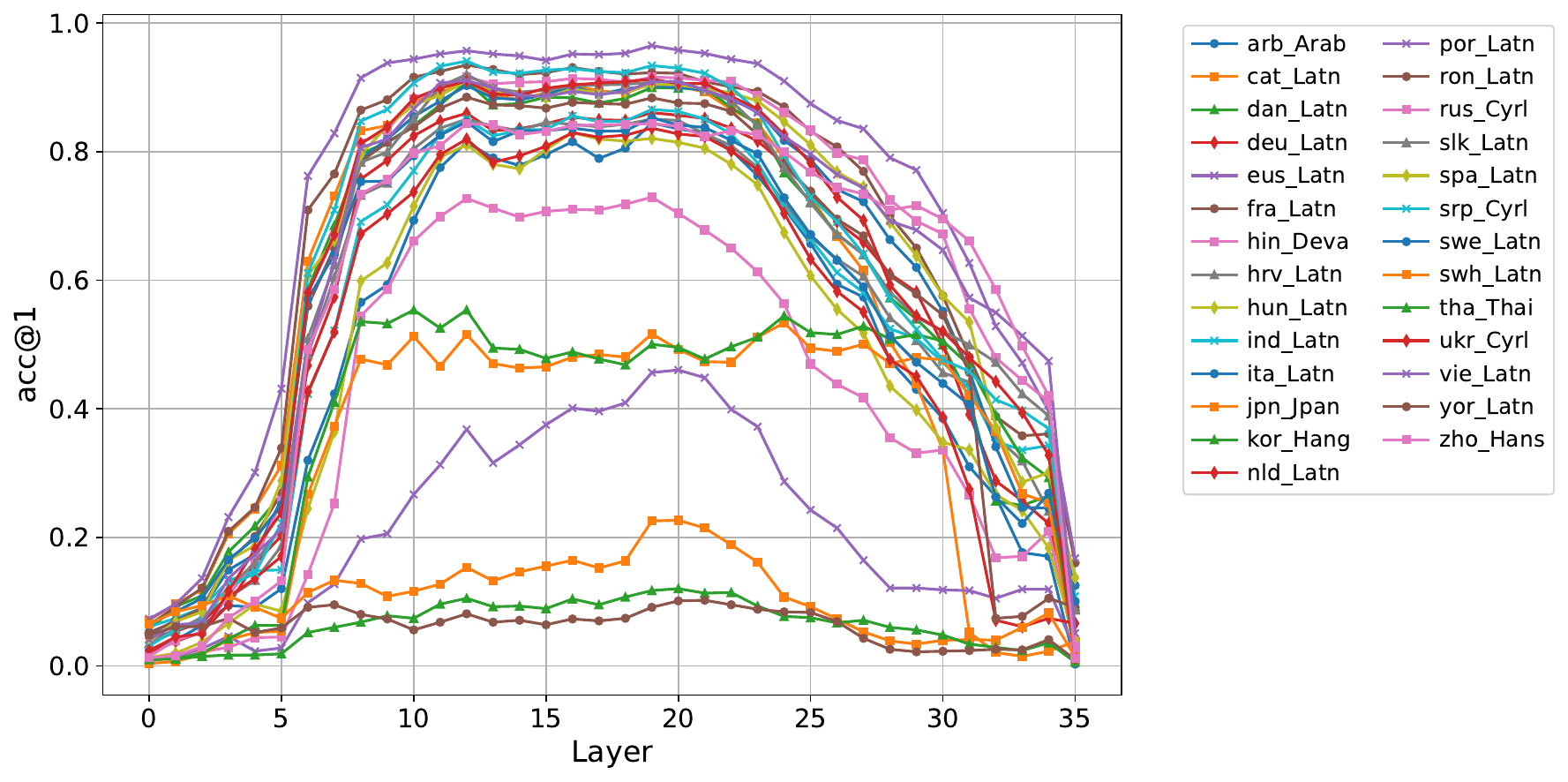}
    \caption{Random}
\end{subfigure}

\vspace{0.5em}

\begin{subfigure}{0.9\linewidth}
    \centering
    \includegraphics[width=\linewidth]{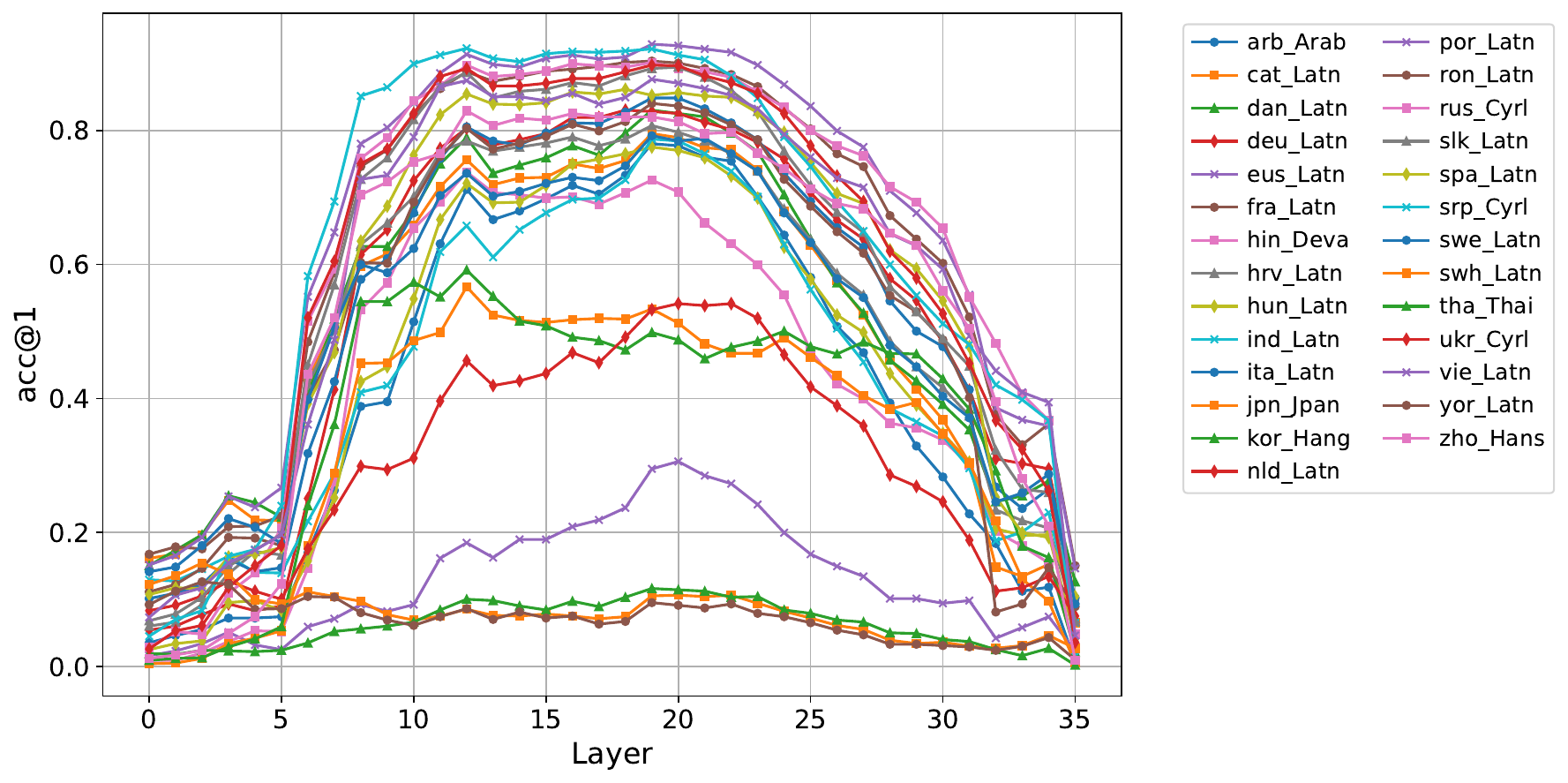}
    \caption{Character}
\end{subfigure}

\caption{Retrieval experiments on FLORES-200 for Qwen}
\label{fig:retrieval-qwen}
\end{figure*}

\begin{figure*}[t!]
\centering

\begin{subfigure}{0.9\linewidth}
    \centering
    \includegraphics[width=\linewidth]{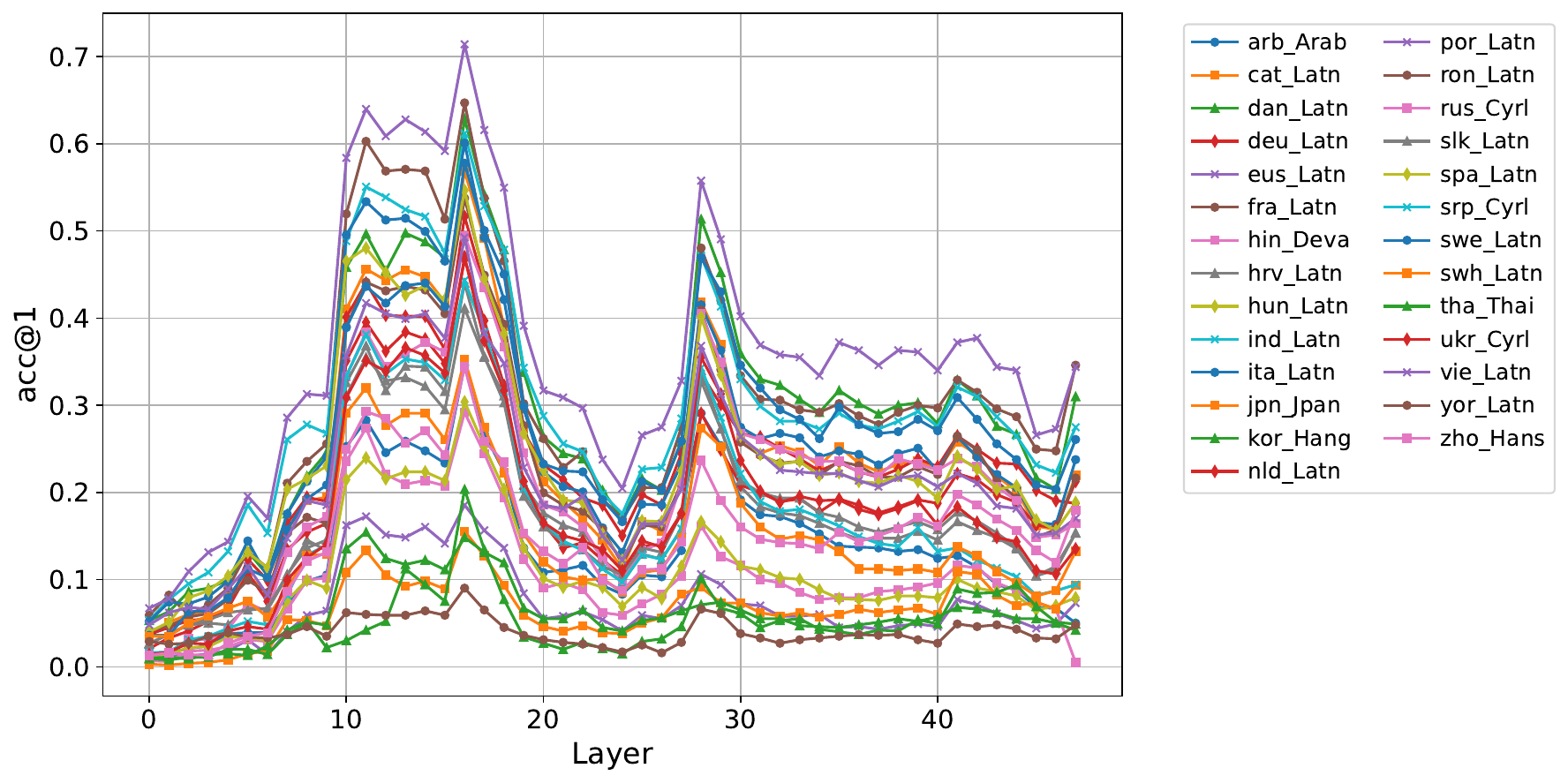}
    \caption{Normal}
\end{subfigure}

\vspace{0.5em}

\begin{subfigure}{0.9\linewidth}
    \centering
    \includegraphics[width=\linewidth]{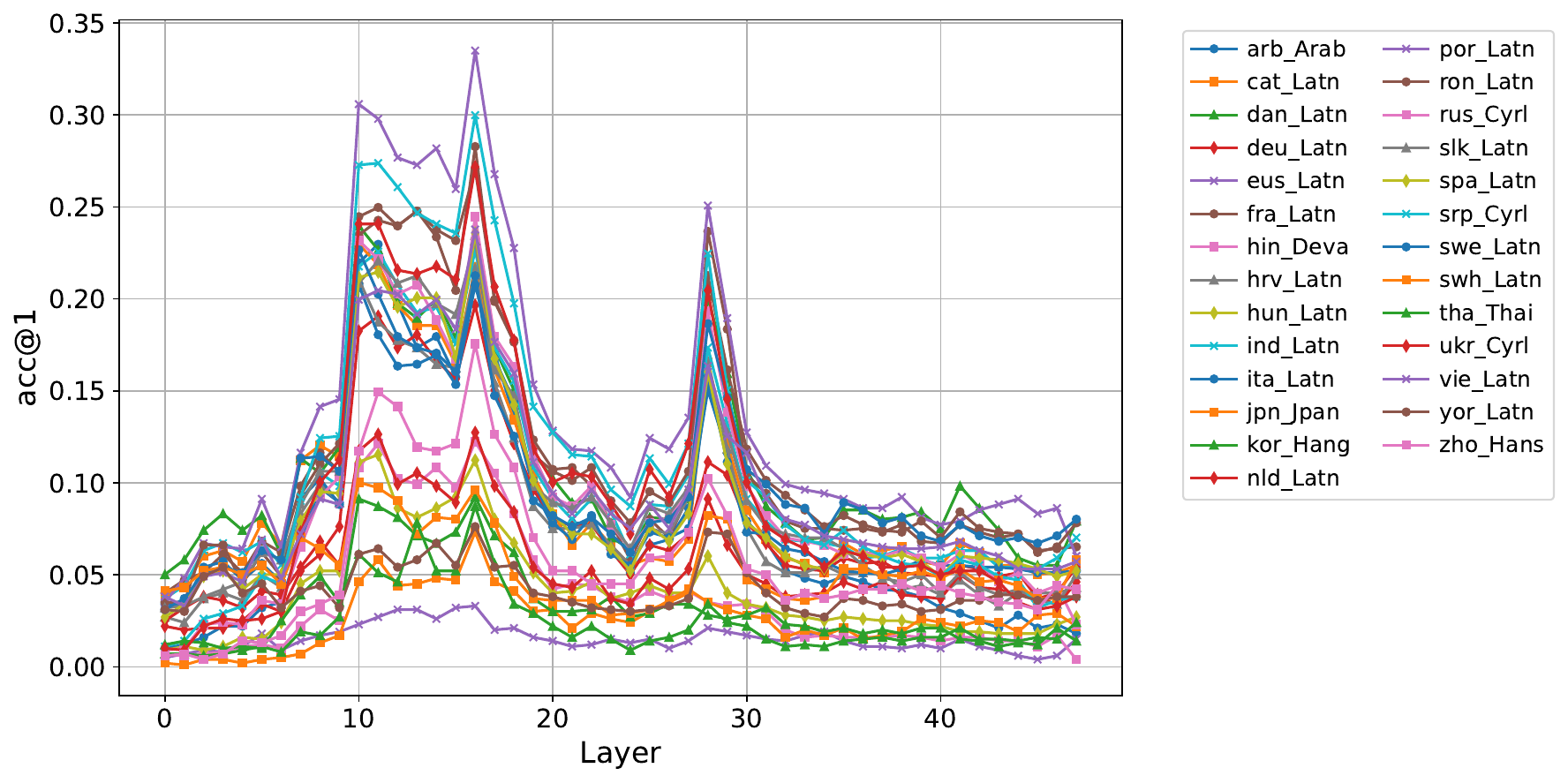}
    \caption{Random}
\end{subfigure}

\vspace{0.5em}

\begin{subfigure}{0.9\linewidth}
    \centering
    \includegraphics[width=\linewidth]{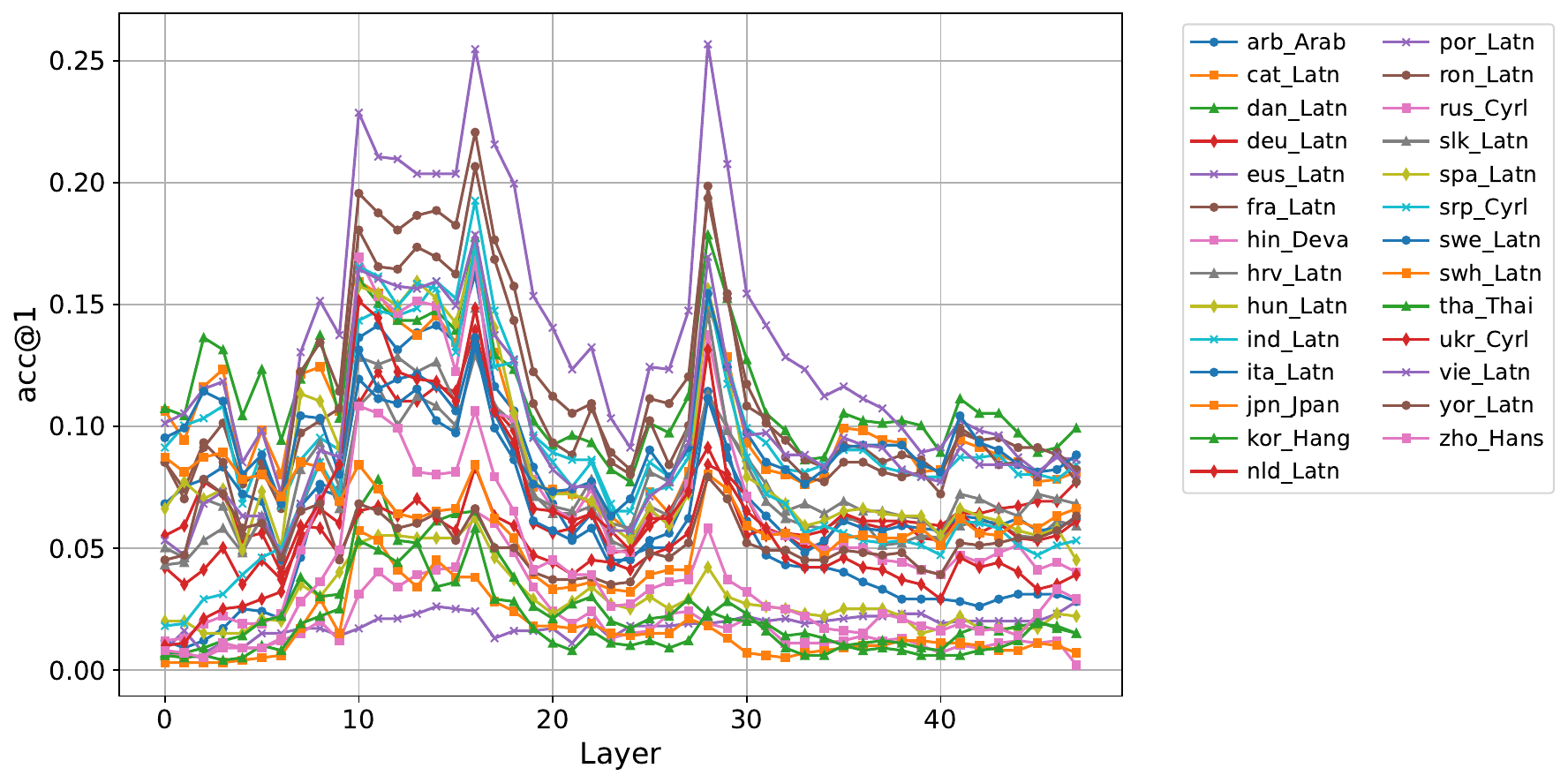}
    \caption{Character}
\end{subfigure}

\caption{Retrieval experiments on FLORES-200 for Gemma}
\label{fig:retrieval-gemma}
\end{figure*}

\begin{table*}[t]
\centering
\small
\begin{tabular}{lllll}
\hline
\textbf{Language} & \textbf{ISO-639-1} & \textbf{Language Tag} & \textbf{Script} & \textbf{Language Family} \\
\hline
Arabic & ar & arb\_Arab & Arabic & Afro-Asiatic (Semitic) \\
Catalan & ca & cat\_Latn & Latin & Indo-European (Romance) \\
Danish & da & dan\_Latn & Latin & Indo-European (Germanic) \\
German & de & deu\_Latn & Latin & Indo-European (Germanic) \\
English & en & eng\_Latn & Latin & Indo-European (Germanic) \\
Basque & eu & eus\_Latn & Latin & Language Isolate \\
French & fr & fra\_Latn & Latin & Indo-European (Romance) \\
Hindi & hi & hin\_Deva & Devanagari & Indo-European (Indo-Aryan) \\
Croatian & hr & hrv\_Latn & Latin & Indo-European (Slavic) \\
Hungarian & hu & hun\_Latn & Latin & Uralic \\
Indonesian & id & ind\_Latn & Latin & Austronesian \\
Italian & it & ita\_Latn & Latin & Indo-European (Romance) \\
Japanese & ja & jpn\_Jpan & Japanese & Japonic \\
Korean & ko & kor\_Hang & Hangul & Koreanic \\
Dutch & nl & nld\_Latn & Latin & Indo-European (Germanic) \\
Portuguese & pt & por\_Latn & Latin & Indo-European (Romance) \\
Romanian & ro & ron\_Latn & Latin & Indo-European (Romance) \\
Russian & ru & rus\_Cyrl & Cyrillic & Indo-European (Slavic) \\
Slovak & sk & slk\_Latn & Latin & Indo-European (Slavic) \\
Spanish & es & spa\_Latn & Latin & Indo-European (Romance) \\
Serbian & sr & srp\_Cyrl & Cyrillic & Indo-European (Slavic) \\
Swedish & sv & swe\_Latn & Latin & Indo-European (Germanic) \\
Swahili & sw & swh\_Latn & Latin & Niger-Congo (Bantu) \\
Thai & th & tha\_Thai & Thai & Kra-Dai \\
Ukrainian & uk & ukr\_Cyrl & Cyrillic & Indo-European (Slavic) \\
Vietnamese & vi & vie\_Latn & Latin & Austroasiatic \\
Chinese (Simplified) & zh & zho\_Hans & Simplified Chinese & Sino-Tibetan \\
\hline
\end{tabular}
\caption{Languages}
\label{tab:languages}
\end{table*}

\begin{table*}[t]
\centering
\small
\begin{tabular}{lcc}
\toprule
\textbf{Language} & \textbf{Canonical Accuracy Coefficient} & \textbf{Canonical Fragmentation Coefficient} \\
\midrule
English (\texttt{en})      & 3.416 & -1.114 \\
Basque (\texttt{eu})       & 2.211 & -0.138 \\
Italian (\texttt{it})      & 2.692 & -0.146 \\
Chinese (\texttt{zh})      & 4.718 & -0.967 \\
\bottomrule
\end{tabular}
\caption{
Logistic regression coefficients on Multilingual-ARC predicting non-canonical performance from canonical performance and canonical fragmentation rate (FR). Positive coefficients for canonical performance indicate that examples solved under canonical tokenization are substantially more likely to remain correct under non-canonical tokenizations. Negative coefficients for canonical fragmentation rate indicate that more fragmented canonical tokenizations are associated with reduced robustness to non-canonical tokenizations.
}
\label{tab:multilingual_arc_logit}
\end{table*}

\begin{figure*}[t!]
\centering

\begin{subfigure}{0.9\linewidth}
    \centering
    \includegraphics[width=\linewidth]{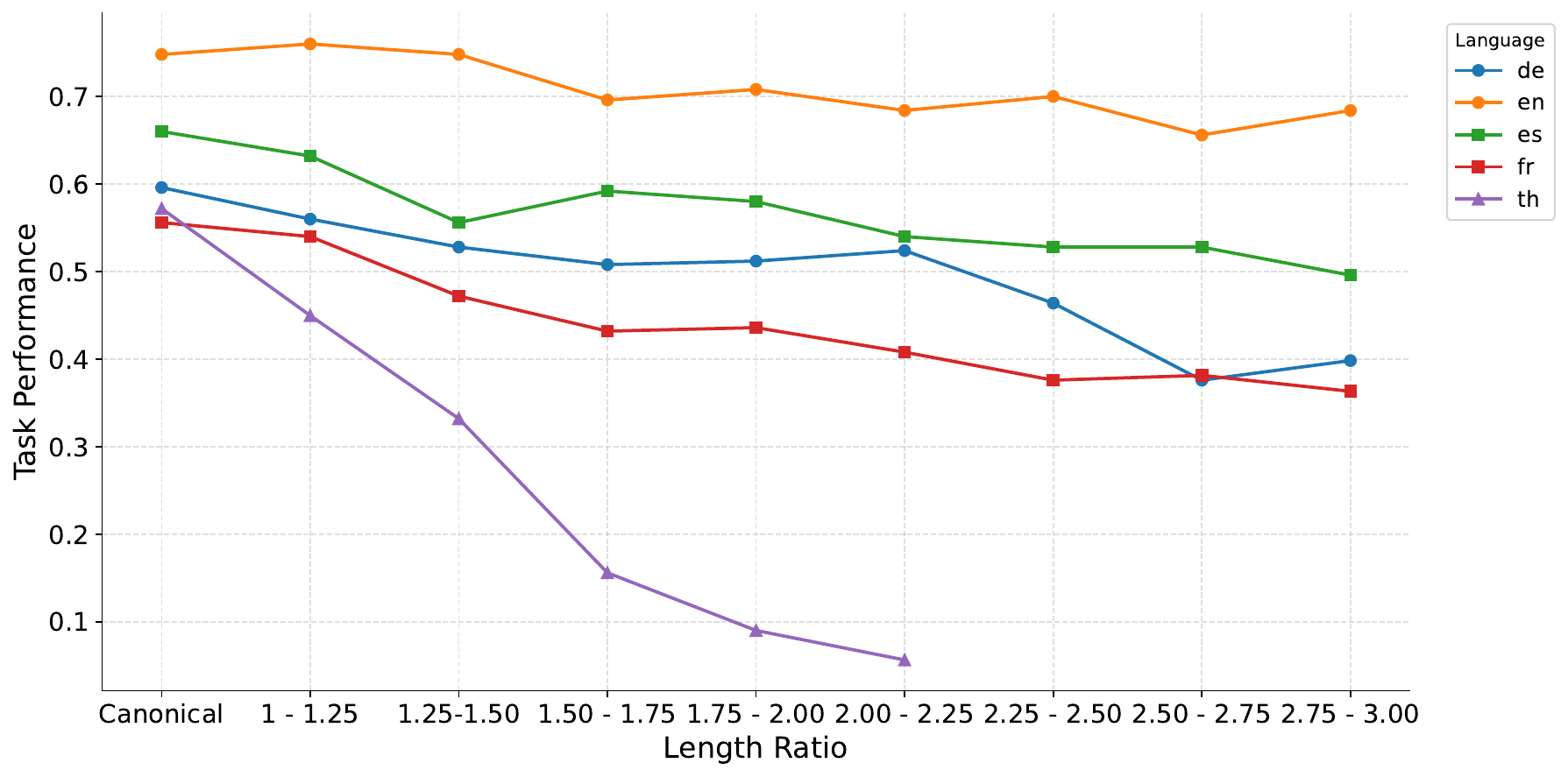}
    \caption{Llama}
\end{subfigure}

\vspace{0.5em}

\begin{subfigure}{0.9\linewidth}
    \centering
    \includegraphics[width=\linewidth]{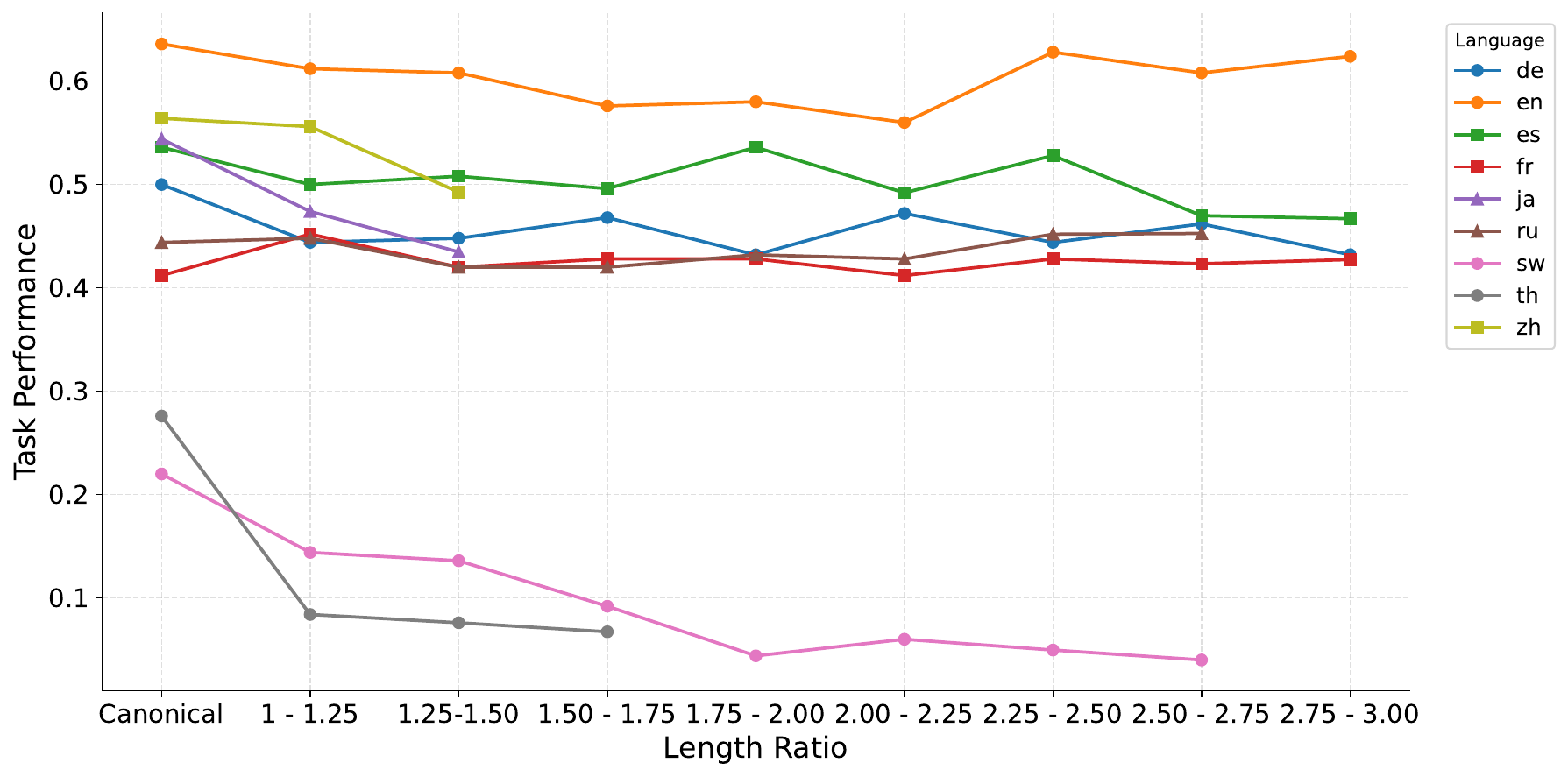}
    \caption{Qwen}
\end{subfigure}

\vspace{0.5em}

\begin{subfigure}{0.9\linewidth}
    \centering
    \includegraphics[width=\linewidth]{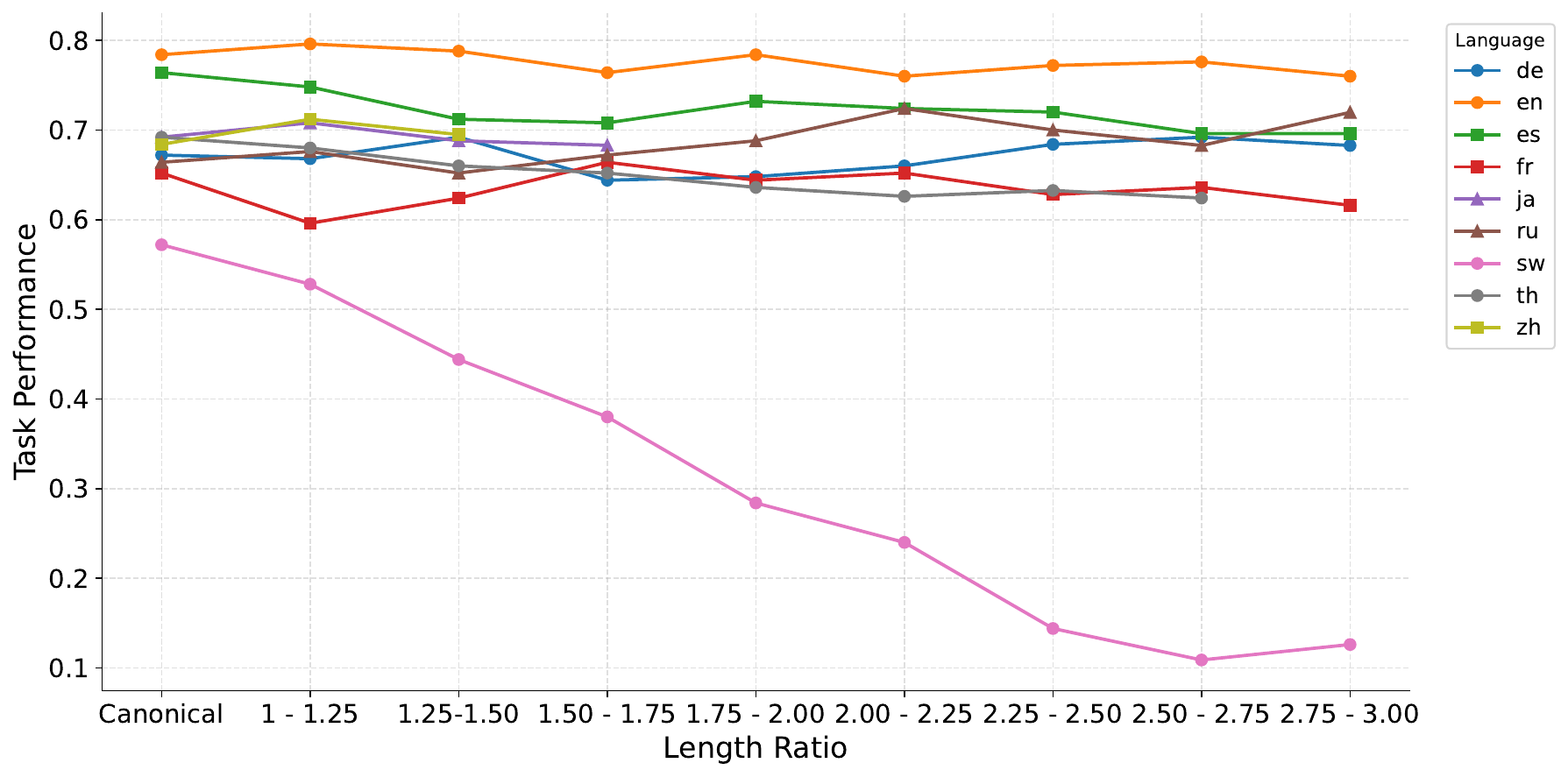}
    \caption{Gemma}
\end{subfigure}

\caption{Impact of granularity of non-canonical tokenization on MGSM performance.}
\label{fig:granularity_mgsm}
\end{figure*}

\begin{figure*}[t!]
\centering

\begin{subfigure}{0.6\linewidth}
    \centering
    \includegraphics[width=\linewidth]{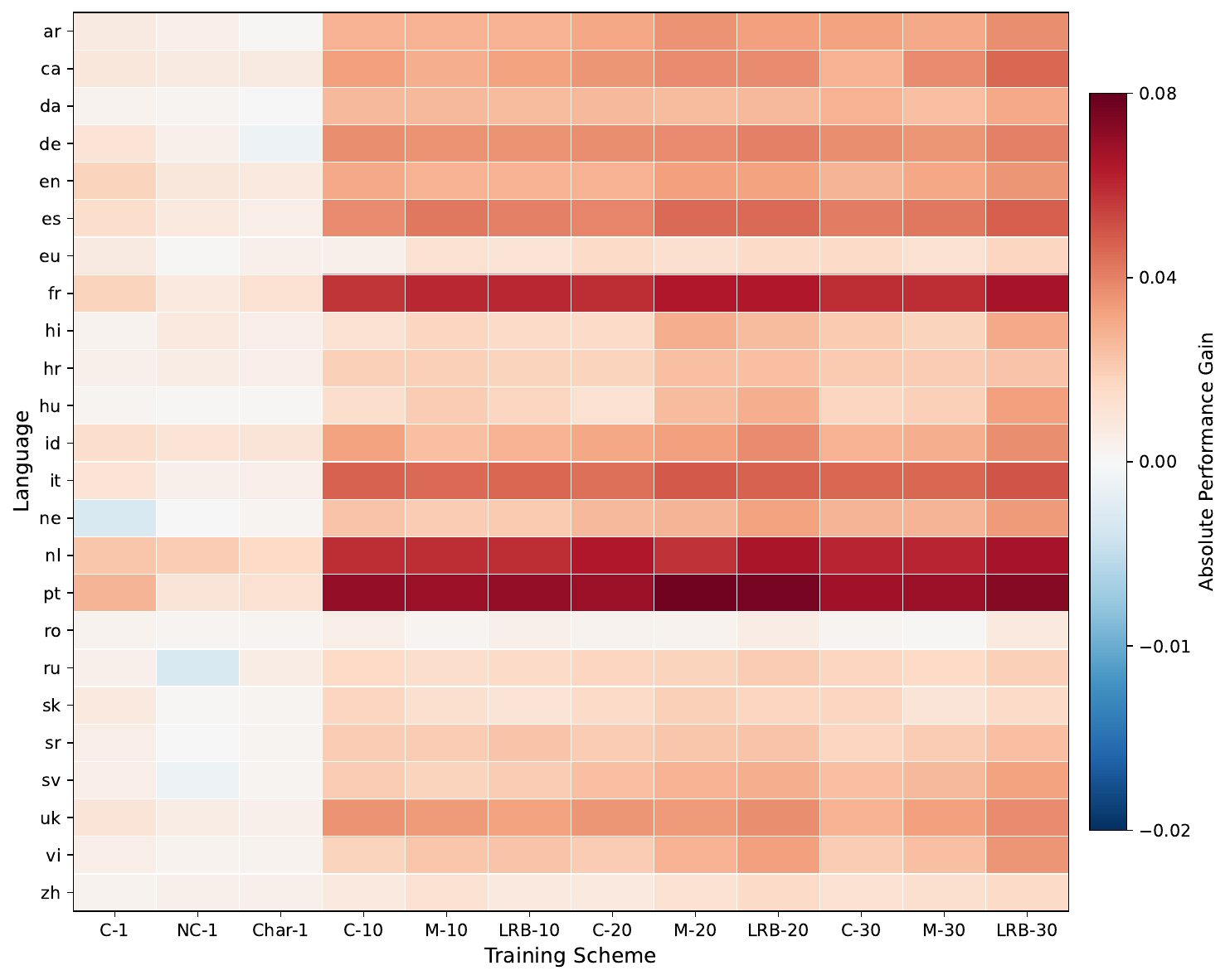}
    \caption{Canonical}
\end{subfigure}

\vspace{0.1em}

\begin{subfigure}{0.6\linewidth}
    \centering
    \includegraphics[width=\linewidth]{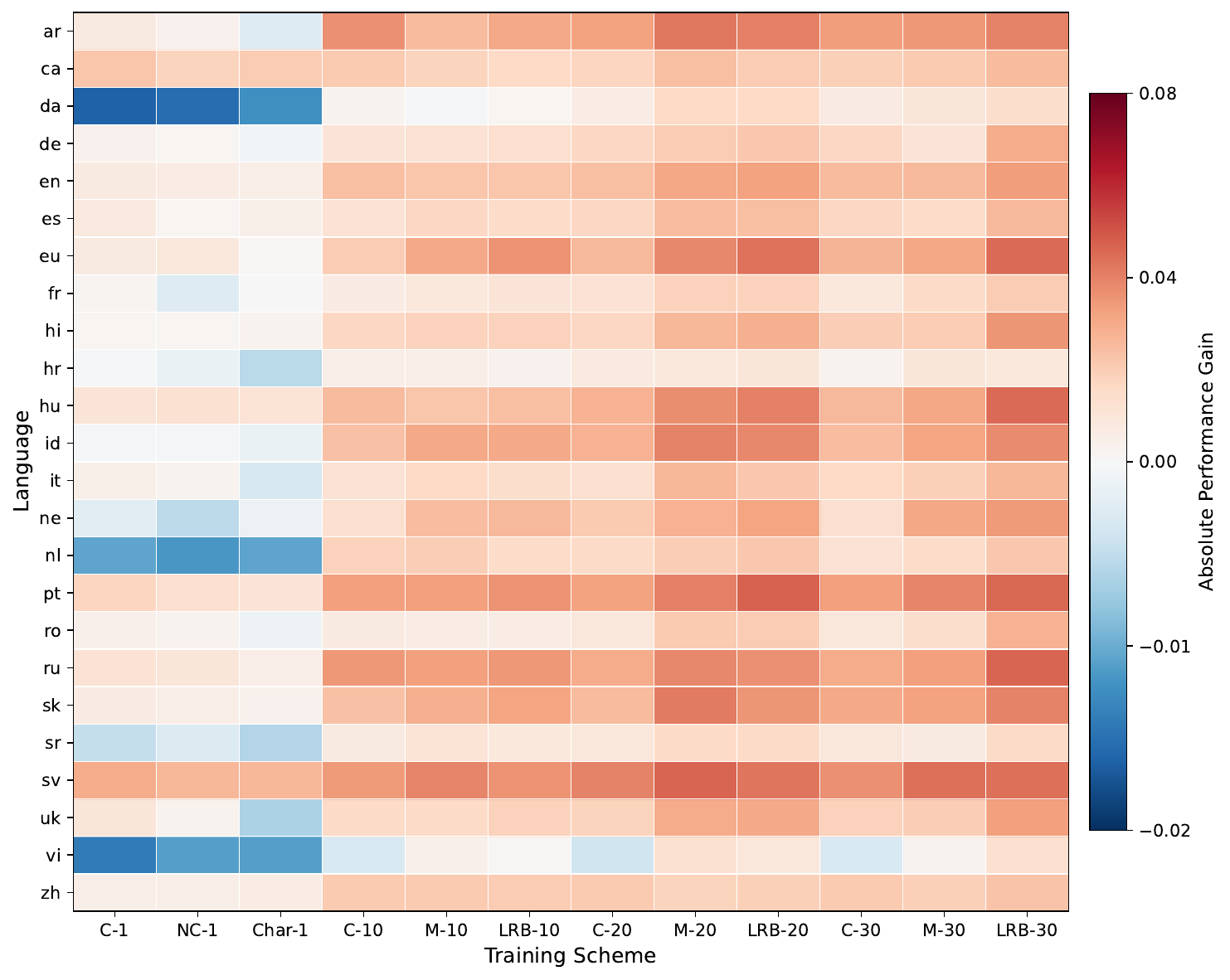}
    \caption{Random Non-canonical}
\end{subfigure}

\vspace{0.1em}

\begin{subfigure}{0.6\linewidth}
    \centering
    \includegraphics[width=\linewidth]{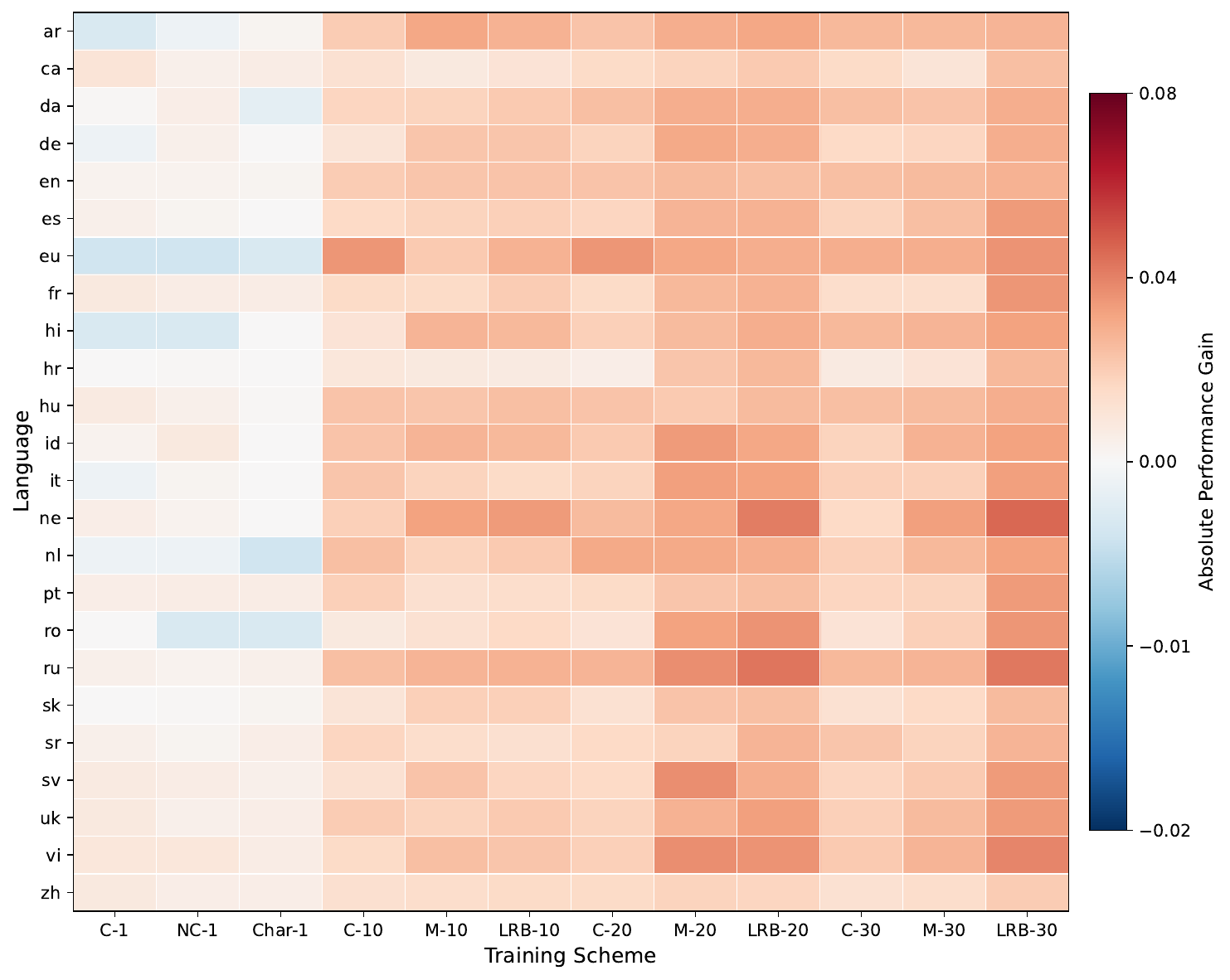}
    \caption{Character}
\end{subfigure}

\caption{Impact of LoRA fine-tuning across languages.}
\label{fig:lora_languages}
\end{figure*}

\end{document}